\documentclass{article}

\usepackage[table,xcdraw]{xcolor} % 在导言区加载 xcolor 包

\usepackage{PRIMEarxiv}
\usepackage{amsmath} 
\usepackage{graphicx}
\usepackage{hyperref}
\usepackage{nicematrix,booktabs}
\usepackage{caption}
\usepackage[utf8]{inputenc} % allow utf-8 input
\usepackage[T1]{fontenc}    % use 8-bit T1 fonts
\usepackage{hyperref}       % hyperlinks
\usepackage{url}            % simple URL typesetting
\usepackage{booktabs}       % professional-quality tables
\usepackage{amsfonts}       % blackboard math symbols
\usepackage{nicefrac}       % compact symbols for 1/2, etc.
\usepackage{microtype}      % microtypography
\usepackage{lipsum}
\usepackage{fancyhdr}       % header
\usepackage{graphicx}       % graphics
\usepackage{float} 
\usepackage{comment}
\usepackage[edges]{forest}
\usepackage{graphicx}
\usepackage{adjustbox}
\usepackage{xcolor}
\usepackage{tikz}
\usepackage{amsmath}
\usepackage{makecell}
\usepackage{booktabs}
\usepackage{graphicx}
\usepackage{booktabs}
\usepackage{multirow}
\usepackage{array}
\definecolor{hidden-draw}{RGB}{106,142,189}
\definecolor{hidden-blue}{RGB}{194,232,247}
\definecolor{hidden-orange}{RGB}{217, 232, 252}
\usepackage{colortbl}
\usepackage{arydshln}
\graphicspath{{media/}}     % organize your images and other figures under media/ folder
\newcommand{\tick}{\checkmark} % Define tick symbol
\newcommand{\cross}{$\times$} % Define cross symbol

\usepackage{array}
\usepackage{graphicx} % for \resizebox
\usepackage{hyperref} % for \cite
\title{Efficiently Integrate Large Language Models with Visual Perception: A Survey from the Training Paradigm Perspective}
\author{
Xiaorui Ma \\
  School of Data Science\\
  Lingnan University \\
  Hong Kong\\
  \texttt{xiaoruima@ln.hk} \\
   \And
  Haoran Xie \thanks{Corresponding author}\\
  School of Data Science\\
  Lingnan University \\
  Hong Kong \\
  \texttt{hrxie@ln.edu.hk} \\
     \And
  S. Joe Qin \\
  School of Data Science\\
  Lingnan University \\
  Hong Kong \\
  \texttt{joeqin@ln.edu.hk} \\
}

\begin{document}
\maketitle

\begin{abstract}
The integration of vision-language modalities has been a significant focus in multimodal learning, traditionally relying on Vision-Language Pretrained Models. However, with the advent of Large Language Models (LLMs), there has been a notable shift towards incorporating LLMs with vision modalities. Following this, the training paradigms for incorporating vision modalities into LLMs have evolved. Initially, the approach was to integrate the modalities through pretraining the modality integrator,  named Single-stage Tuning. It has since branched out into methods focusing on performance enhancement, denoted as Two-stage Tuning, and those prioritizing parameter efficiency, referred to as Direct Adaptation. However,  existing surveys primarily address the latest Vision Large Language Models (VLLMs) with Two-stage Tuning, leaving a gap in understanding the evolution of training paradigms and their unique parameter-efficient considerations. This paper categorizes and reviews 34 VLLMs from top conferences, journals, and highly cited Arxiv papers, focusing on parameter efficiency during adaptation from the training paradigm perspective. We first introduce the architecture of LLMs and parameter-efficient learning methods, followed by a discussion on vision encoders and a comprehensive taxonomy of modality integrators. We then review three training paradigms and their efficiency considerations, summarizing benchmarks in the VLLM field.  To gain deeper insights into their effectiveness in parameter efficiency, we compare and discuss the experimental results of representative models, among which the experiment of the Direct Adaptation paradigm is replicated. Providing insights into recent developments and practical uses, this survey is a vital guide for researchers and practitioners navigating the efficient integration of vision modalities into LLMs.

\end{abstract}

% keywords can be removed

\section{Introduction}
\begin{figure}[H]
            \centering
            \includegraphics[width=1\linewidth]{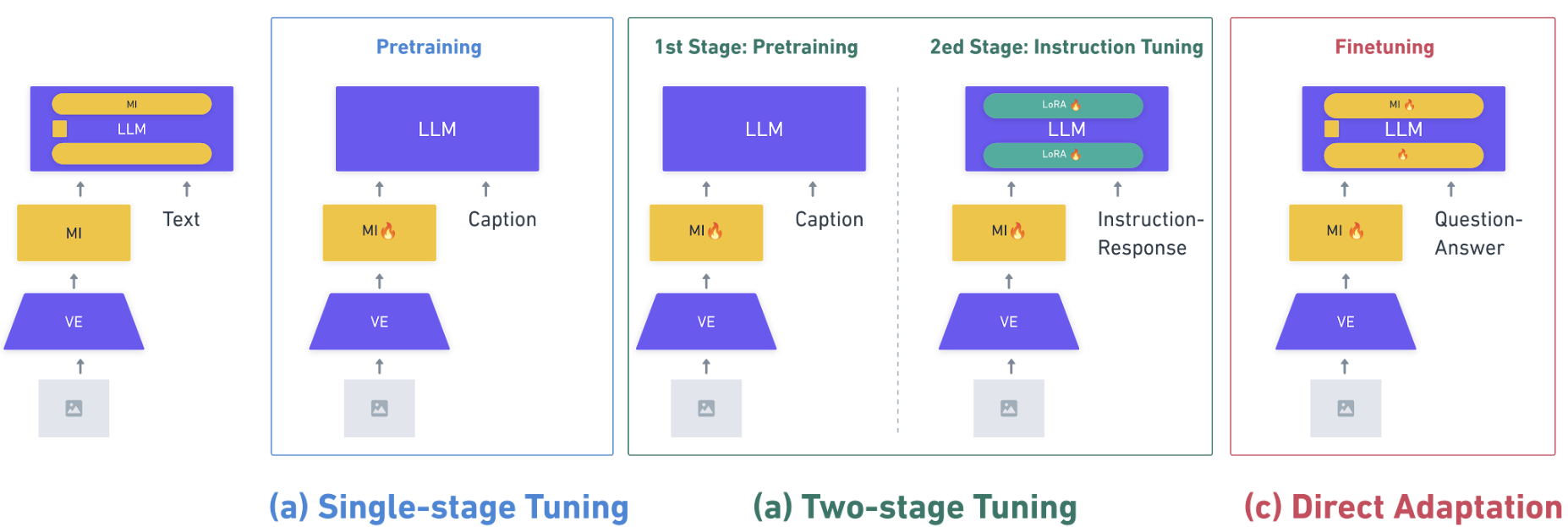}
            \caption{\textbf{Integrated Modules and Three Training Paradigms.} MI denotes modality integrator, and VE denotes vision encoder. The trainable module and learning paradigm are the most adopted.}

            \label{fig: Training Paradigm}
            
        \end{figure}

The study of vision-language modalities has long been a significant topic, with numerous works dedicated to utilizing transformer-based models to perform multimodal learning \cite{du2022survey,yin2024survey}. In the era of Large Language Models (LLMs), multimodal-to-text generation tasks have experienced a paradigm shift from Vision-Language Pretrained Models (VLPMs) \cite{awadalla2023openflamingo,li2022blip,yu2022coca} to integrating LLMs with vision modalities \cite{tsimpoukelli2021multimodal,li2023blip,liu2024visual,gao2023llama}. This shift is driven by the advantages of LLMs in terms of adaptability and reasoning ability. VLPMs require per-task fine-tuning to transfer to downstream tasks, while LLMs have strong zero-shot or few-shot adaptation abilities \cite{tsimpoukelli2021multimodal}, saving the resources needed for per-task tuning. In addition, although VLPMs have visual perception abilities, enabling them to identify and caption objects in an image, they lack reasoning capabilities \cite{yang2023enhance}.
In contrast, LLMs can leverage their pretrained knowledge to reason with visual information \cite{lu2022learn, liu2023mmbench, fu2023mme}, offering a deeper understanding of images. While LLMs have these advantages, leveraging off-the-shelf LLMs for VLPM is challenging due to their integrated architecture \cite{li2023blip}, where the vision encoder and text encoder are constituted as a single model. In contrast, adding a vision encoder to an LLM is more straightforward, requiring a Modality Integrator (MI) to connect the two models. The resulting model is named as Vision Large Language Models (VLLMs), and the architecture is shown in Figure \ref{fig: Training Paradigm}.
As LLMs scale, computational resource demands increase, making parameter efficiency critical in building VLLMs \cite{jin2024efficient, xu2023parameter}. This survey examines the Parameter-Efficient Adaptation (PEA) techniques for incorporating visual modalities into LLMs from the training paradigm perspective. The training paradigms are categorized into three types: Single-stage Tuning, Two-stage Tuning, and Direct Adaptation. The categorization is driven by the fact that each paradigm has distinct motivations for efficiency, and different methods are employed to achieve it.
\\
VLLMs adopting Single-stage Tuning first appeared in the VLPM era. From the parameter efficiency perspective, pretraining a VLPM requires multiple feedforward processes due to the simultaneous use of various learning objectives \cite{du2022survey}, resulting in the trainable parameters increasing multiplicatively as the model size increases. By adding LLMs with visual perception through a Single-stage Tuning paradigm, in most cases, only an MI is trained to bridge two modalities in one training process \cite{li2023blip, alayrac2022flamingo, mokady2021clipcap, manas2022mapl, najdenkoska2023meta, koh2023grounding}. Compared to LLM’s scale, this is also a parameter-efficient strategy. For example, BLIP-2 \cite{li2023blip} utilizes Flan-T5-XXL with 11 billion parameters, while MI accounts for 0.89\% of the whole model. For downstream tasks generalization, unlike VLPMs that adopt end-to-end per-task fine-tuning, zero-shot, and few-shot learning are adopted in Single-stage Tuning to leverage the pretrained knowledge in LLMs.
\\
However, Single-stage Tuning cannot fully unlock the generalization potential and instruction-following capabilities of LLMs. For better zero-shot transfer to unseen tasks and user intentions understanding, Two-stage Tuning introduces an additional training phase, instruction tuning, that involves fully training LLMs in the second stage \cite{liu2024visual}. 
Due to the large size of LLMs, there are three methods to reduce trainable parameters: not training LLM but only the MI in the second stage \cite{li2023blip, zhu2023minigpt, wang2023cogvlm, yang2022zero, zhang2023video, alayrac2022flamingo}, training the MI while incorporating reparameterization modules into LLMs \cite{liu2024visual, liu2024improved, bai2023qwen, ye2024mplug, ye2023mplug} such as LoRA \cite{hu2021lora}, and utilizing a smaller LLMs \cite{chu2023mobilevlm, chu2024mobilevlm, qiao2024vl}.
\\
In contrast to Two-stage Tuning aiming to improve VL performance, Direct Adaptation primarily focuses on consuming the least resources to transfer LLM to the VL domain. It skips the pretraining stage and directly finetunes the MI on downstream tasks mainly through multi-task learning without updating LLMs \cite{sung2022vl,shukor2023ep, liang2022modular, jie2024memory, zhang2023llama, sung2022lst, liang2024querying, hu2023vl, luo2024cheap}. The design of MI achieves an excellent balance between parameter efficiency and modality fusion performance.

\begin{figure}[H]

                \centering
                 
        \includegraphics[width=1\linewidth]{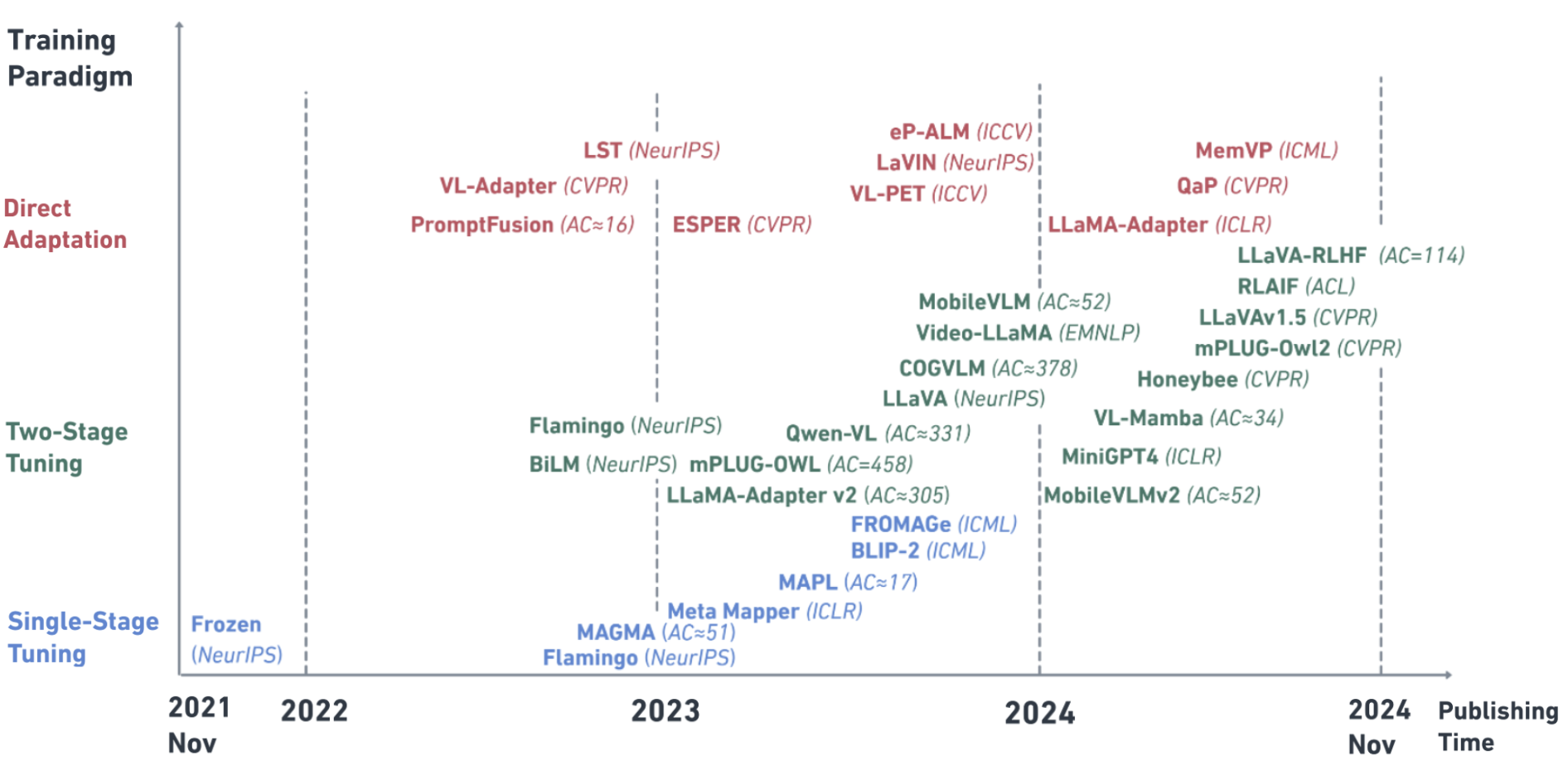}
        \caption{\textbf{The Taxonomy and Publishing Time.} AC denotes the annual citation times. For published work, the horizontal axis shows the Published time, while for unpublished work, it shows the submission time to Arxiv.}

                \label{fig:Publishing Time}

\end{figure}

Existing surveys \cite{jin2024efficient,yin2024survey}, however, mainly focus on the latest VLLMs adopting Two-stage Tuning paradigms. 
In this survey, the database used is Google Scholar \cite{GoogleScholar}, and the keywords are Multimodal, Large Language model, vision-language model, and parameter-efficient learning. The time period is from November 2021 to November 2024. The search results are first screened to match the idea that the model integrates vision modality into LLMs and considers parameter efficiency. Then, more papers are included from the related work of the screened literature. Finally, the quality of the papers is assessed. The inclusion criteria are that the paper needs to be published in the conference ranking A and B in the CCF Recommended List of International Conferences and Periodicals \cite{CCF}, or ICLR. If it does not satisfy the former requirement or it’s an Arxiv paper, the annual citation should exceed 15 times as of November 1st, 2024. Based on this selection process, this review surveys 34 papers on this topic, which are shown in Figure \ref{fig:Publishing Time}. The details of the reviewed models are presented in the Table \ref{tab: Summary}. 
\\
In addition, this review is arranged based on the steps to integrate an LLM with visual perception: 
\begin{enumerate}
    \item An LLM is selected as the base model to be augmented with the new modality. (See Section \ref{sec: LLM}) This section introduces the transformer architecture, along with existing PEA methods.
    \item A vision encoder is chosen to encode the image into a hidden representation. (See Section \ref{sec: VE}) This section comprehensively summarizes the architecture and pretrained modalities of employed vision encoders.
    \item An MI is designed to transform the visual embedding to the semantic space of LLM. (See Section \ref{sec: Modality_Int}) In this section, the two main categories of modality integrators, Out-of-block Integrators and In-block Integrators, are elaborated.
    \item A training paradigm is adopted to transfer LLM to the VL domain.  (See Section \ref{sec: training paradigm}) Three training paradigms, Single-stage Tuning, Two-stage Tuning, and Direct Adaptation, are discussed, with a detailed exploration of the corresponding training techniques, datasets, parameter-efficient strategies, and performance evaluations for each approach.
\end{enumerate}
Finally, we conclude the paper in Section \ref{sec: future direction}.

\begin{table*}[!htb]
\fontsize{9.2pt}{8.1pt}\selectfont
\centering
\renewcommand{\arraystretch}{2} % 增加行距

\resizebox{\textwidth}{!}{

\begin{tabular}{p{3cm} p{1.8cm} p{1.8cm} p{2.5cm} p{4cm} p{2cm} p{1.8cm} p{1.8cm} p{2.5cm} p{1cm} p{1cm}}

\toprule
\textbf{Model} & \textbf{LLM} & \textbf{LLM Size} & \textbf{Vision Encoder} & \textbf{Modality Integrator} & \textbf{Visual Input Form} & \textbf{PT} & \textbf{FT} & \textbf{Learning Paradigm} & \textbf{PT size} & \textbf{FT size} \\

\midrule
\textbf{Single Stage Tuning} & & & & & & & & & & \\

\underline{Frozen} \cite{tsimpoukelli2021multimodal} & GPT2-like & 7B & NF-ResNet-50 & Linear Projector & Soft Prompt & MI + VE & - & - & - & - \\

ClipCap \cite{mokady2021clipcap} & GPT-2 & 1.5B & CLIP ViT-B/32 & Resampler & Soft Prompt & MI & - & - & 3M & - \\
MAGMA \cite{eichenberg2021magma} & GPT-J & 6B & CLIP-ResNet-50x16 & Linear Projector + Bottleneck Adapter$^\dagger$ & Soft Prompt & MI + VE & - & - & 25M & - \\
\underline{Flamingo} \cite{alayrac2022flamingo} & Chinchilla & 1.4B/7B/70B  & NF-ResNet-F6 & Resampler + Attention-based Adapter$^\dagger$ & Prefix & MI & - & - & 2.2B & - \\
MAPL \cite{manas2022mapl} & GPT-J & 6B & CLIP ViT-L/14 & Resampler & Soft Prompt & MI & - & -& 398K & - \\
ESPER \cite{ESPER2023Yu} & GPT-2-base &117M  &CLIP ViT-B/32 &MLP 2x &Soft Prompt & MI &- &RL &- &- \\
FROMAGe \cite{koh2023grounding} & OPT & 6.7B & CLIP ViT-L/14 & Linear Projector & Soft Prompt & MI & - & - & 3.3M & - \\
MetaMapper \cite{najdenkoska2023meta} & GPT-2 & 1.5B & CLIP ViT-B/32 & Resampler & Prefix & MI & - & - & 330K & - \\
BLIP-2 \cite{li2023blip} & OPT/FlanT5 & 2.7B/6.7B 3B/11B & ViT-L/14/ ViT-g/14 & Q-Former + Linear Projector & Soft Prompt & MI & - & - & 129M & - \\

\midrule
\textbf{Two-stage Tuning} & & & & & & & & & & \\
\underline{Flamingo} \cite{alayrac2022flamingo}\cite{awadalla2023openflamingo} &Chinchilla & 1.4B/7B/70B & NF-ResNet-F6 & Resampler + Attention-based Adapter$^\dagger$ & Prefix & MI & MI & Per-task & 2.2B & - \\
BiLM \cite{yang2022zero} & DeBERTa & 900M & CLIP ViT-L/14 & Linear Projector + Bottleneck Adapter$^\dagger$ & Soft Prompt & MI & MI & Per-task & 10M & - \\
LLaMA-Adapter \cite{zhang2023llama} & LLaMA1 & 7B & ViT-B/16 & Linear Projector + Degree-Adaptive Prefix$^\dagger$ & Prefix & MI & Partial MI & Instruction (T+V) & - & - \\
ESPER \cite{ESPER2023Yu} & GPT-2-base &117M  &CLIP ViT-B/32 &MLP 2x &Soft Prompt & MI &MI &RL + Per-task &- &- \\
LLaVA \cite{liu2024visual} & LLaMA1 & 7B & CLIP ViT-L/14 & Linear Projector & Soft Prompt & MI & MI + LLM$^\ddagger$ & Instruction (V) & 558K & 158K \\
MiniGPT4 \cite{zhu2023minigpt} & Vicuna & 13B & EVA-CLIP ViT-G & Linear Projector + Q-former & Soft Prompt & Partial MI  & Partial MI  & Instruction (V) & 5M & 5K \\
mPLUG-Owl1 \cite{ye2023mplug} & LLaMA1 & 7B & CLIP ViT-L/14 & Resampler & Soft Prompt & MI + VE & MI + LoRA & Instruction (T+V) & 2.1M & 102K \\
LLaMA-Adapter v2  \cite{gao2023llama} & LLaMA1 & 7B/65B & CLIP ViT-L/14 & Linear Projector + Degree-Adaptive Prefix$^\dagger$ & Prefix & MI & Bias, Norm & Instruction (T+V) & 567K & 52K \\
Video-LLaMA \cite{zhang2023video} & LLaMA1/ Vicuna & 7B/13B & EVA-CLIP ViT-G/14 & Q-Former + Linear Projector & Soft Prompt & MI & MI & Instruction (V) & 2M & 244K \\
BLIP-2 \cite{li2023blip} & OPT/FlanT5 & 2.7B/6.7B 3B/11B & ViT-L/14/ ViT-g/14 & Q-Former + Linear Projector & Soft Prompt & MI & MI & Per-task & 129M & - \\
QWEN-VL \cite{bai2023qwen} & Qwen & 7B & OpenCLIP ViT-bigG-14 & Resampler & Soft Prompt & MI + VE & MI + LLM$^\ddagger$ & Multi-task + Instruction (T+V) & 1.4B & 50M \\
LLaVA-RLHF \cite{sun2024llavaRLHF} & Vicuna-v1.5  & 7B/13B & CLIP ViT-L/14 & MLP 2x & Soft Prompt & MI+LoRA & MI + LoRA &Instruction (V)+RL &-  &- \\
LLaVA 1.5 \cite{liu2024improved} & Vicuna-v1.5  & 7B/13B & CLIP ViT-L/14 & MLP 2x & Soft Prompt & MI & MI + LLM$^\ddagger$ &Instruction (V) & 558K & 665K\\
CogVLM-Chat \cite{wang2023cogvlm} & Vicuna-v1.5 & 7B & EVA02-CLIP-E/14 & MLP2x + Unimodal Linear Adapter$^\dagger$& Prefix & MI & MI+VE & Multi-task + Instruction (V) & 1.5B & 6M \\
mPLUG-Owl 2 \cite{ye2024mplug} & LLaMA2 & 7B & CLIP ViT-L/14 & Resampler + Unimodal Linear Adapter$^\dagger$ & Soft Prompt & MI + VE & MI + VE + LLM$^\ddagger$ & Multi-task + Instruction (T+V) & 348M & 1.23M \\
C-Abstractor \cite{cha2024honeybee} & Vicuna-v1.5 & 7B/13B & CLIP ViT-L/14 & Convolution-based Abstractor& Soft Prompt & MI & MI + LLM & Multi-task + Instruction (V) & 200M & 8.13M \\
D-Abstractor \cite{cha2024honeybee} & Vicuna-v1.5 & 7B/13B & CLIP ViT-L/14 & Attention-based Abstractor & Soft Prompt & MI & MI + LLM & Multi-task  + Instruction (V) & 200M & 8.13M \\
MobileVLM \cite{chu2023mobilevlm} & MobileLLaMA & {1.4B/2.7B} & CLIP ViT-L/14 & Convolution-based Abstractor & Soft Prompt & MI & MI + LLM$^\ddagger$ & Instruction (V) & 558K & 665K \\
MobileVLM v2 \cite{chu2024mobilevlm} & MobileLLaMA & {1.4B/2.7B}& CLIP ViT-L/14 & Convolution-based Abstractor & Soft Prompt & MI + LLM & MI + LLM$^\ddagger$ & Multi-task + Instruction (V) & 1.2M & 2.4M \\
\underline{VL-Mamba} \cite{qiao2024vl} & Mamba LLM &{2.8B} & CLIP ViT-L/14 & VSS-based Abstractor & Soft Prompt & MI & MI + LLM & Instruction (V) & 558K & 665K \\
RLAIF \cite{RLAIF2024Ahn} & Vicuna-v1.5 &7B/13B &CLIP ViT-L/14 & MLP 2x &Soft Prompt  & MI + LoRA & MI + QLoRA & Instruction (V) + RL & 20K &40k \\

\midrule
\textbf{Direct Adaptation} & & & & & & & & & & \\
VL-Adapter \cite{sung2022vl} & $\text{BART}$/ $\text{T5}$ & 139M/220M & CLIP-ResNet-101 & Linear Projector + Bottleneck Adapter$^\dagger$ & Soft Prompt & - & MI & Multi-task  & - & - \\
\underline{PromptFusion} \cite{liang2022modular} & GPT3 & 175B & NF-ResNet-50 & - & Soft Prompt & - & Soft Prompt & - & - & - \\
LST \cite{sung2022lst} &T5 &220M &CLIP-ResNet-101 &Linear Projector + Attention-based Adapter$^\dagger$ &Soft Prompt & - & MI &Multi-task  &- \\
eP-ALM \cite{shukor2023ep} & OPT & 2.7B & ViT-B/16 & Linear Projector& Prefix & - & MI & Per-task  & - & - \\

LaVIN \cite{luo2024cheap} & LLaMA1 & 7B/13B & CLIP ViT-L/14 & {MLP2x + Bottleneck Adapter$^\dagger$} & Soft Prompt & - & MI & Instruction (T+V) & - & - \\
VL-PET \cite{hu2023vl} & $\text{BART}$/ $\text{T5}$ & 139M/220M & CLIP ViT-L/14 & Linear Projector + Bottleneck Adapter$^\dagger$ & Soft Prompt & - & MI & Multi-task  & - & - \\
MemVP \cite{jie2024memory} & Bart/ T5/ LLaMA1 & 7B/13B & CLIP ResNet-101 & Linear Projector& Prefix & - & MI & Per-task & - & - \\
QaP \cite{liang2024querying} & DeBERTa & 900M & CLIP ViT-L/14 & Attention-based Adapter$^\dagger$ + Bottleneck Adapter$^\dagger$ + Degree-Adaptive Prefix$^\dagger$ & Prefix & - & MI & Per-task  & - & - \\

\bottomrule

\end{tabular}
}
\caption{ \textbf{Summary of Reviewed Models.} MI and VE refer to the Modality Integrator and Vision Encoder. The underlined models are closed-source models, and in open-sourced models, the code of QaP \cite{liang2024querying} and x-LLM \cite{chen2023x} have not yet been released. The PT and FT sizes of BLIP2, Mini-GPT4, mPLUG-Owl, Qwen-VL, and LLaVA v1.5 are from \cite{qiao2024vl}. $^\dagger$ denotes the In-block Integrators. $^\ddagger$ denotes that the work provides LoRA-based Tuning in the implementation. Per-task, Multi-task, and Instruction denote Per-task Finetuning, Multi-task Finetuning, and Instruction Tuning. RL denotes reinforcement learning. T and V denote the modality of Instruction Tuning data. The sequence of papers in the table follows the time submitted to Arxiv for consistent comparison with the initial idea. }
\label{tab: Summary}
\end{table*}

\section{Large Language Model}
\label{sec: LLM}
Large Language Models (MML) are mainly transformer-based \cite{vaswani2017attention} models with encoder \cite{liu2019roberta,he2020deberta}, encoder-decoder \cite{raffel2020exploring,lewis2020bart} and decoder-only \cite{touvron2023llama,Vicuna,zhang2022opt,floridi2020gpt,achiam2023gpt,bai2023qwen,glm2024chatglm} architectures. LLMs utilized in the reviewed paper are summarized in Table \ref{tab: LLMs}. In the context of the three training paradigms discussed in this study, the selection of LLMs varies distinctively. Overall, the LLM release time is the key factor for the choice. For the Single-stage Tuning paradigm, the GPT series is typically adopted as this paradigm is early.
In the Two-Stage Tuning paradigm, 7B and 13B LLaMA and Vicuna models are generally utilized. Exceptions are that MobileVLM \cite{chu2023mobilevlm, chu2024mobilevlm} , and VLMamba \cite{gu2023mamba} leverage smaller-scale LLMs to enhance efficiency.
In the Direct Adaptation paradigm, T5 \cite{raffel2020exploring} and Bart \cite{chung2024scaling} are employed as the benchmark LLM due to their manageable sizes \cite{sung2022vl, sung2022lst,hu2023vl, luo2024cheap, jie2024memory}. Additionally, there is a notable trend towards adapting LLaMA models in this paradigm \cite{luo2024cheap,jie2024memory,zhang2023llama,gao2023llama}.

As the preliminaries, the transformer architecture and existing PEA methods for LLMs are introduced in this section.

\begin{table*}[!htb]
\small
\centering

\resizebox{\textwidth}{!}{
\begin{tabular}{lllll}

\toprule
 \textbf{LLM} &  \textbf{Instruction Tuned} &  \textbf{Architecture} &  \textbf{Release Time} &  \textbf{Available Size}  \\ 
 \midrule
        GPT-2 \cite{brown2020language} &-  & Decoder-Only & Feb 2019 & 117M/345M/774M/1.5B  \\ 
        RoBERTa \cite{liu2019roberta} &-  & Encoder-only & Jul 2019 & 125M/355M  \\ 
        T5 \cite{raffel2020exploring} & Flan T5 \cite{chung2024scaling} & Encoder-Decoder & Oct 2019/ Oct 2022 & 220M/770M/3B/11B  \\ 
        BART \cite{chung2024scaling} &-  & Encoder-Decoder &  Oct 2019 & 139M  \\ 
        DeBERTa V2 \cite{he2020deberta} &-  & Encoder-only & Jun 2020 & 900M/1.5B  \\ 
        GPT-J \cite{gpt-j} &-  & Decoder-Only & May 2021 & 6B  \\ 
        Chinchilla \cite{hoffmann2022training} &- &Decoder-Only  & Mar 2022 & 70B \\
        OPT  \cite{zhang2022opt} &-  & Decoder-Only & May 2022 &\multirow{2}{*}{\makecell[l]{125M/350M/1.3B/2.7B/ \\ 6.7B/13B/30B/66B/175B}}    \\ 
        \\
        LLaMA1 \cite{touvron2023llama} & Vicuna  \cite{Vicuna} & Decoder-Only & Feb 2023/ Mar 2023 & 7B/13B/33B/65B  \\ 
       
        MPT  \cite{MPT2023} & MPT-Instruct & Decoder-Only & May 2023 & 1B/7B/30B  \\ 
        RedPajama \cite{redpajama} & RedPajama -Instruct & Decoder-Only & May 2023 & 3B/7B  \\ 
        LLaMA2 \cite{touvron2023llama2} & Vicuna v1.5 & Decoder-Only & July 2023/ Aug 2023 & 7B/13B/34B/70B  \\ 
        Qwen \cite{bai2023qwen} &-  & Decoder-Only & Sep 2023 & 1.8B/7B/14B/72B  \\ 
        MobileLLaMA \cite{chu2023mobilevlm} &-  & Decoder-Only &  Dec 2023 & 1.4B/2.7B  \\ 
        Mamba \cite{gu2023mamba} &-  & Selective State Space Models & Dec 2023 & 130M/370M/790M/1.4B/2.8B  \\ 
\bottomrule
\end{tabular}
}
\caption{\textbf{Summary of LLMs.} }
\label{tab: LLMs}
\end{table*}

\begin{table}[!htb]
\small  % 设置字体大小为 12pt，行间距为 14pt

\centering

\begin{tabular}{llll}

\toprule
\textbf{Notation}              & \textbf{Description}                     \\ 

\midrule
\textbf{Matrix}              &                                     &                               \\
$I, T$                                & The raw input to the MLLM, i.e. image, text.                              \\
$X, X_v, X_t $                            & The embedding output by vision encoder or tokenizer.             \\
$P', P_v', P_t'$ & The random-initialized soft prompt or prefix.  \\
$P,P_v,P_t$                            & The output soft prompt or prefix.                          \\
$S,S_v, S_t$                            & The output of the attention module in the transformer block.                     \\
$h_i^t $                        & The output of $i$-th transformer block and $t$-th timestep.               \\

$\hat{Y}$                        & The output of LLM.                        \\

$Y$                            & The ground truth Label.                            \\

\midrule
\textbf{Size}                &                                        \\
$L$                            & The number of transformer blocks in an Attention-based Integrator or LLM.                 \\
$n,n_v,n_t $                           & The length of embedding.                         \\
$d,d_v,d_t $                            & The dimension of embedding.                 \\
$r$                            & The low rank.                      \\ &                 &                  \\

\midrule
\textbf{Parameter}           &                                     &                 &                  \\
$\phi$                          & The frozen parameters, usually refer to the frozen LLMs' parameters.                    &            \\
$\theta $                       & The trainable parameters.                             \\

\midrule
\textbf{Network}           &                                                \\
$W_i(\cdot)$ & The $i_\text{th}$ linear layer in an MLP. & &\\
$W_{down}(\cdot) $                     & The down projection layer that reduces the dimension of the input.
              \\
$W_{up}(\cdot)$                        & The up projection layer that increases the dimension of the input.
                            \\
$W^Q(\cdot),W_v^Q(\cdot),W_t^Q(\cdot)$ & The linear layer that transforms the input into query vectors.  \\
$W^K(\cdot),W_v^K(\cdot),W_t^K(\cdot)$ & The linear layer that transforms the input into key vectors.  \\
$W^V(\cdot),W_v^V(\cdot),W_t^V(\cdot)$ & The linear layer that transforms the input into value vectors.
 \\
$W^O(\cdot)$ & The linear layer that transforms the concatenated heads into outputs. \\
$W_\text{FFN}(\cdot) $     & The feed-forward module in a transformer block in equation \ref{eq:FFN}.                               \\
$\text{VE}(\cdot)$                         & The vision encoder that extracts visual features from the image.
                              \\
$\text{LLM}(\cdot)$                        & The large language model.                       \\
$\text{Block}(\cdot) $                     & A transformer block in the Attention-based Integrator or LLM.
       
          \\
${W}_{\text{Conv}_i}(\cdot)$ & The $i$-th convolution layer. \\
${W}_{\text{PatchWise}}(\cdot)$ & The point-wise convolution layer that convolutes with a kernel size of 1.
                          \\
${W}_{\text{DepthWise}}(\cdot) $                        & The depth-wise convolution layer that convolutes independently over each channel.
           \\  
             
${W}_{\text{BatchNorm}}(\cdot) $                         & The batch normalization layer.                           \\
${W}_{\text{LayerNorm}}(\cdot) $                         & The layer normalization layer.                     \\
$\text{ResBlock}_i(\cdot)$ & The $i$ -th residual block in equation \ref{eq:conv}. \\ 
$\text{CLIP-T}(\cdot)$ & The text encoder of CLIP. \\

\midrule
\textbf{Operation}           &                                   \\
${[};{]} $                     & The concatenation.                                \\ 
$\text{Attention}(\cdot)$ &The attention mechanism in equation \ref{eq:attn}. \\
$\text{Softmax}(\cdot)$  &The softmax operation.  \\ 
$\sigma (\cdot)$ &The activation function. \\ 

\bottomrule
\end{tabular}

\caption{\textbf{The Mathematical Notations.}}
\label{tab:notations}
\end{table}

\subsection{Transformer}
\label{sec: transformer}
Transformer \cite{vaswani2017attention} comprises an encoder and a decoder, each of which includes $L$ transformer blocks. The basic modules for a transformer block are Multi-Head Attention (MHA) and the Feed-forward Network (FFN). After each module, there is a Layer Normalization (LN) and a residual connection.  
For a self-attended MHA, the input embedding $X$ is linearly transformed $h$ times and activated by the attention mechanism to get the $\text{head}$. The concatenated heads will then be projected to the model dimension $d$.

\begin{equation}
\begin{aligned}
Q = X {W^Q},\quad K = X{W^K},\quad V = X{W^V}, \\
\text{Attention}(Q,K,V) = \text{softmax}\left(QK^T / \sqrt{D}\right)V, \\
\text{MultiHead}(Q, K, V) = [\text{head}_1; \ldots; \text{head}_h)] \cdot W^O, \\
\text{where } \text{head}_i = \text{Attention}(QW_i^Q, KW_i^K, VW_i^V),
\end{aligned}
\label{eq:attn}
\end{equation}

and the parameter size of the linear layers are $ W_i^Q \in \mathbb{R}^{d \times d_q}$ , $W_i^K \in \mathbb{R}^{d \times d_k}$ $, W_i^V \in \mathbb{R}^{d} \times d_v $ $ W^O \in \mathbb{R}^{hd_v \times d}$. The mathematical notations are summarized in the Table \ref{tab:notations}.  In cross-attention, the queries are derived from the target sequence, while the keys and values come from a context sequence, which is different from self-attention, where all values are from the same input.
The FFN module is an MLP composed of two linear layers and activated by RELU.
\begin{equation}
\text{FFN}(X) =  \sigma( X W_1 ) W_2,
\label{eq:FFN}
\end{equation}

where $\sigma(x)=max(0,x)$.

Based on the transformer architecture, LLaMA 1 \cite{touvron2023llama} improves LM to RMSNorm, changing ReLU activation into SwiGLU activation \cite{shazeer2020glu}, further enhancing the model's nonlinear expression. LLaMA 2 \cite{touvron2023llama2} replaces MHA with Grouped-Query Attention, where KV projections are shared across groups of heads instead of all heads.

\subsection{Parameter-Efficient Adaptation}
\label{sec: PEA}

PEA is a solution that transfers LLMs to new tasks by updating a small number of parameters \cite{xu2023parameter}. Typically, there are three categories of methods: Prompt-based Tuning \cite{li2021prefix, lester2021power}, Adapter-based Tuning \cite{houlsby2019parameter}, LoRA-based Tuning \cite{hu2021lora,qlora2023Dettmers,liu2024dora}. In the multimodal context, these ideas are inherited in a way that the visual feature is fed into the LLMs in the form of visual prompts or prefixes, following which the visual modality is fused with textual modality by adapters or LoRA.

\subsubsection{Prompt-based Tuning}
The prompt-based method is to add trainable embeddings to the LLM input. The prompt-based method can be divided into prefix tuning and soft prompt tuning, which differs in that the soft prompt tuning only adds trainable vectors at the very first input. In contrast, the prefix tuning adds trainable queries to the input for each transformer layer. In the multimodal context, soft prompts and prefixes are the forms of visual embedding input into the LLM.

\paragraph{Prefix Tuning}
The prefix tuning \cite{li2021prefix} adds a trainable prefix to the input layer and each transformer layer. As the subsequent generation is conditioned on the prefix,  it can serve as a learnable instruction for different downstream tasks. For the timesteps within the prefix length $p$, the transformer layer output is the updated prefix embedding $P$, and for the later timesteps, which are the textual inputs, the output is from the frozen language model but conditioned on $P$. Instead of directly updating the prefix embedding $P'_\theta$, an MLP and a smaller matrix ($P'$) are used for parameterization, thus ensuring the training stability. The output of the $t$-th timestep can be formulated as:

\begin{equation}
\begin{aligned}
    h^t &= 
    \begin{cases} 
    P[t, :] = \text{MLP}_\theta(P'[t, :]), & \text{if } t \in p, \\
    \text{LM}_\phi(\left[ P; X \right], h^{<t}), & \text{otherwise}.
    \end{cases} \\
    h^t &= \left[ h_1^{t} ; \ldots ; h_n^{t} \right], \\
\end{aligned}
\end{equation}

where $P_\theta \in p  \times d$ and $h^t$ are the concatenation of the output of all transformer layers at timestep $t$.

\paragraph{Soft Prompt Tuning}
Lester et al. \cite{lester2021power} simplifies the prefix tuning idea to add a trainable prompt only to the input of the language model as a learnable instruction to guide the model to perform different downstream tasks. The model is now maximizing $\Pr_{\phi; \theta}(Y \mid [P; X])$, where $\theta$ is the soft prompt parameter and $\phi$ is the frozen LLM's parameter, and the shape of the input matrix is that $[P; X] \in \mathbb{R}^{(p+n) \times d}
$.

In addition to the textual prompt tuning, Visual Prompt Tuning (VPT) \cite{jia2022visual} proposes the visual version of prompt tuning (VPT- shallow) and prefix tuning (VPT- deep), which is to add learnable vector in the input layer and in each transformer layer respectively. For both scenarios, the input to the first transformer encoder layer is $[X_{[CLS]};P;X_v]$, where the soft prompt is added after the $[CLS]$ token. For multimodality context, PromptFuse \cite{liang2022modular} directly adds a soft prompt at the beginning of concatenated visual embedding $X_v$ and text embedding $X_t$. The input can be formulated as $[P;X_v;X_t]$.
The idea of integrating vision modality into the language model by inputting image-conditioned soft prompt originates from Frozen \cite{tsimpoukelli2021multimodal}. To avoid hurting the LM's generalization ability by a relatively small amount of multimodal training data, the LM is kept frozen, and the vision encoder and a linear layer are trainable to align the two modalities. In this way, Frozen only trains 0.56\% of the total model parameters.
The difference between soft prompt in LLM and in VLLMis that the former learns the difference between downstream tasks, while the visual soft prompt represents not the difference, but the image-conditioned information.

\subsubsection{Adapter-based Tuning}
\label{sec: Bottleneck Adapter}
Adapter-based tuning \cite{houlsby2019parameter} is to insert a trainable parameter-efficient module into the transformer architecture. In NLP, the architecture of adapters is usually the bottleneck structure \cite{houlsby2019parameter}, while different works propose different inserting positions \cite{lin2020exploring,he2021towards} and training strategies \cite{chronopoulou2023adaptersoup}. It projects down and up the output matrices over the $d$ dimension, which can be formulated as:
\begin{equation}
\begin{aligned} 
    &X = (\text{ReLU}(X   W_{\text{down}}))   W_{\text{up}} + X, \\
    &W_{\text{down}} \in \mathbb{R}^{d \times r}, W_{\text{up}} \in \mathbb{R}^{r \times d}, \text{where } r<<d.
\end{aligned}
\end{equation}

In a multimodal context, the In-block Modality Integrator inherits the idea of adding efficient trainable modules into LLM, but there are various structures in addition to the bottleneck structure. More details are discussed in Section \ref{sec: In-block Integrator}.

\subsubsection{LoRA-based Tuning}
\label{sec: LoRA}
LoRA-based tuning is utilized in VLLMs involving updating the LLMs in the second training stage. LoRA \cite{hu2021lora} updates the transformer weights by decomposing the changing weights to two low-rank matrices, which can be formulated as:
\begin{equation}
    \Delta W=BA, 
    W=W_0+\Delta W,
\end{equation}
where $ B \in \mathbb{R}^{d \times r}, A \in \mathbb{R}^{r \times k},  r \ll \min(d, k)$.
During training, only parameters in the low-rank matrices $B$ and $A$ are updated, keeping the pre-trained parameters $W_0$ frozen. A set of parameters in $B$ and $A$ can be stored for each downstream task. QLoRA \cite{qlora2023Dettmers} advances the LoRA approach by incorporating weight quantization for the LoRA adapters, reducing them to lower precision. This enhancement significantly decreases both memory usage and storage needs. Decomposed Rank Adaptation (DoRA) \cite{liu2024dora} further refines the process of model fine-tuning by decomposing the pretrained weights into two components: magnitude and direction. By leveraging LoRA to fine-tune the directional component efficiently, DoRA maintains parameter efficiency while simultaneously avoiding additional inference latency. 

\subsection{Learning Paradigm}
\label{sec: Learning Paradigm}
The learning paradigms in VLLMs are adapted from LLMs, including Multi-task learning (MTL), Instruction Tuning, and Reinforcement learning (RL). This section outlines their use in LLMs and briefly covers their adaptation to multimodal settings. The learning paradigms of each model are summarized in Table \ref{tab: Summary}. 
\subsubsection{Multi-task Learning}
MTL refers to training a model to tackle multiple tasks concurrently during a single training phase \cite{liu2019multi}, which utilizes common knowledge across multiple related tasks for better generalization performance \cite{caruana1997multitask}.  Its complementary approache is per-task learning. MTL has become a key approach in NLP, demonstrating a wide range of applications such as information extraction, natural language understanding, and text generation \cite{zhang2023multitasksurvey}. In the context of LLMs, MTL has evolved to require adjustment of task-specific weights \cite{lai2024multi}. In the multimodal domain, task weights are typically determined by the size of the data for each task \cite{sung2022vl}. The corresponding loss function can be expressed as:

\begin{equation}
    \mathcal{L}(D; \theta) = \frac{1}{|D|} \sum_{(I, T, Y) \in D} l(I, T, Y; \theta),
\end{equation}
where $D$ is the universal VL dataset composed of $N$ datasets $D_1, D_2...D_N$.

\subsubsection{Instruction Tuning}
VLLMs are designed to enable effective communication with humans, and instruction tuning equips VLLMs with the ability to understand user intentions and respond to commands\cite{wei2022finetuned}. Studies have shown that fine-tuning LLMs with instruction data significantly enhances their zero-shot performance on unseen tasks\cite{wei2022finetuned}. The instruction data is built by organizing original structured datasets in different ways connected by natural-language instructions. The instruction data with multimodal information is first proposed by LLaVA \cite{liu2024visual}.

\subsubsection{Reinforcement Learning}
In the LLM context, RL is a learning paradigm where an LLM learns to generate human-preferred outputs by setting the goal of maximizing rewards obtained from human feedback \cite{ouyang2022training}, AI feedback \cite{bai2022constitutional}, or other reward systems \cite{guo2025deepseek}.
During learning, Proximal Policy Optimization (PPO) \cite{schulman2017ppo} is a widely used RL loss function. It limits drastic policy updates through a clipping mechanism, while maximizing the cumulative reward. The standard PPO objective is defined as:
\begin{equation}
\begin{aligned}
     &L^{\text{PPO}}(\theta) = \mathbb{E}_t \left[ \min \left( r_t(\theta) A_t, \text{clip}(r_t(\theta), 1 - \epsilon, 1 + \epsilon) A_t \right) \right], \\
&r_t(\theta) = \pi_\theta(a_t | s_t) \cdot \left[\pi_{\theta_\text{old}}(a_t | s_t)\right]^{-1},
\end{aligned} 
\end{equation}
where $r_t(\theta)$ is the ratio of the new and old policy probabilities, $A_t$ is the advantage function that estimates how much better an action is compared to the expected return, and  $\epsilon$ is a hyperparameter controlling the clipping range. The clipping function ensures that policy updates remain within a constrained range to prevent excessively large updates that destabilize training.

Reinforcement Learning with Human Feedback (RLHF) has become crucial in aligning LLMs with human preferences and ethical considerations \cite{ouyang2022training}. The RLHF pipeline generally consists of three stages. First, in Supervised Fine-tuning (SFT), the LLM is initially tuned by instructions. Second, in Reward Model Training, a separate reward model is trained using human preference on multiple model outputs, ranking them based on quality. Third, the LLM is fine-tuned using the PPO algorithm, backpropagating the LLM parameter. 
In this setting, the model needs to optimize responses based on users' overall preferences rather than every detail. Therefore, lower computational costs are consumed because it updates the model by utilizing coarse-grained feedback, such as paragraph-level ranking or overall preferences, avoiding fine-grained backpropagation for each token \cite{yu2024rlhf}. 
This approach has been pivotal in enhancing LLMs like ChatGPT \cite{ouyang2022training}. In the multi-modal domain, models like LLaVA-RLHF \cite{sun2024llavaRLHF} and RLAIF \cite{RLAIF2024Ahn} adopt a similar training pipeline. Recently, DeepSeek-R1 \cite{guo2025deepseek} has improved efficiency by reducing the number of training examples required in SFT and replacing reward model training through the application of a rule-based reward system. In the multi-modal context, ESPER \cite{ESPER2023Yu} reduces training costs by leveraging unpaired image data and using CLIP similarity as a reward signal.

\section{Vision Encoder}

\label{sec: VE}
To add the visual modality, a pre-trained vision encoder is utilized to extract visual embedding $X_v$ from the input image $I$. The extracted $X_v$ will be further transformed to feed into LLM. The vision encoders used in the reviewed literature are summarized in Table \ref{tab: Vision Encoder}. Overall, CLIP ViT L/14 \cite{radford2021learning} is the most commonly utilized vision encoder, and there is no clear preference in terms of vision encoders for the three training paradigms. 

\begin{table*}[!htb]
\small
\centering

\begin{tabular}{lll}
\toprule
\textbf{Vision Encoders}   & \textbf{Architecture}~~~~~   & \textbf{Parameter Size}    \\ 

\midrule
\textbf{Uni-modal} \\
ViT-B/16 \cite{dosovitskiy2020image} &ViT  & 86.2M\\
ViT-L/14 \cite{dosovitskiy2020image} &ViT  & 304M \\
NF-ResNet-50 \cite{brock2021high} &ResNet  & 25.6M\\
NF-ResNet-F6 \cite{brock2021high} &ResNet  & 438.4M\\
ViT-g/14 \cite{zhai2022scaling} &ViT  & 1.3B \\
\midrule
\textbf{Multimodal} \\
CLIP ViT-L/14 \cite{radford2021learning} &ViT  & 304M                       \\
CLIP ViT-B/32 \cite{radford2021learning}  &ViT    & 87.8M\\
CLIP-ResNet-101 \cite{radford2021learning} &ResNet  & 56.3M                      \\
CLIP-ResNet-50x16 \cite{radford2021learning} &ResNet   & 167.3M \\
OpenCLIP ViT-bigG-14 \cite{cherti2023reproducible} &ViT   & 1.9B                    \\
EVA-CLIP ViT-G/14 \cite{sun2023eva} &ViT     & 1B\\
EVA02-CLIP-E/14  \cite{sun2023eva} &ViT      & 4.4B \\

\bottomrule
\end{tabular} 
\caption{\textbf{Summary of Vision Encoders.} The models are accessed from the OpenClip repository \cite{ilharco_gabriel_2021_5143773}.}
\label{tab: Vision Encoder}
\end{table*}

\subsection{Architecture}
There are two architectures: Vision Transformer (ViT) \cite{dosovitskiy2020image} and Residual Network (ResNet) \cite{he2022deep}, between which ViT is more frequently employed. The core idea of ViT \cite{dosovitskiy2020image}  is to treat an image as a sequence of patches and regard them as tokens, similar to words in a sentence. Assuming that an image $ I \in \mathbb{R}^{h \times w \times c}$, is inputted into the vision encoder, it will first be divided into $N$ patches, where each patch $ I_p \in \mathbb{R}^{a \times a \times c}$ and $N = \frac{hw}{p^2}$. Each patch is flattened into a vector, linearly transformed and added with positional encoding, and then encoded by the transformer encoder, which is described in Sec \ref{sec: transformer}. 

ResNet \cite{he2022deep} is a Convolutional Neural Network (CNN). It is composed of a series of residual blocks, with each block consisting of multiple convolutional layers with skip connections. For a residual block with one convolution layer, the first residual block can be formulated as:
\begin{equation}
    X_v' = \text{ResBlock}(I)= \text{ReLU}(W_{\text{BatchNorm}}(I \cdot W_{\text{Conv}}) +I).
\label{eq:conv}
\end{equation}

\subsection{Pretrained Modality}
From the modality perspective, both uni-modal and multimodal vision encoders are utilized. Uni-modal vision encoder refers to encoders only pre-trained by images, while multimodal vision encoder is the vision encoder of the CLIP model \cite{radford2021learning}. CLIP is a VLPM pre-trained with image-text pairs and contrastive loss, which pushes similar text and visual representations closer while pushing the negative samples further. Most VLLMs adopt a multimodal vision encoder. Merullo et al. \cite{merullo2022linearly} proves that the more language supervision involved pertaining to the image encoder, the better the performance of language vision tasks. Out of data efficiency considerations, eP-ALM \cite{shukor2023ep} choose a uni-modal vision encoder to avoid using multimodal encoders pretrained on huge datasets.

\section{Modality Integrator}
\label{sec: Modality_Int}

\begin{figure}[H]
    \centering
    
    \includegraphics[width=0.8\linewidth]{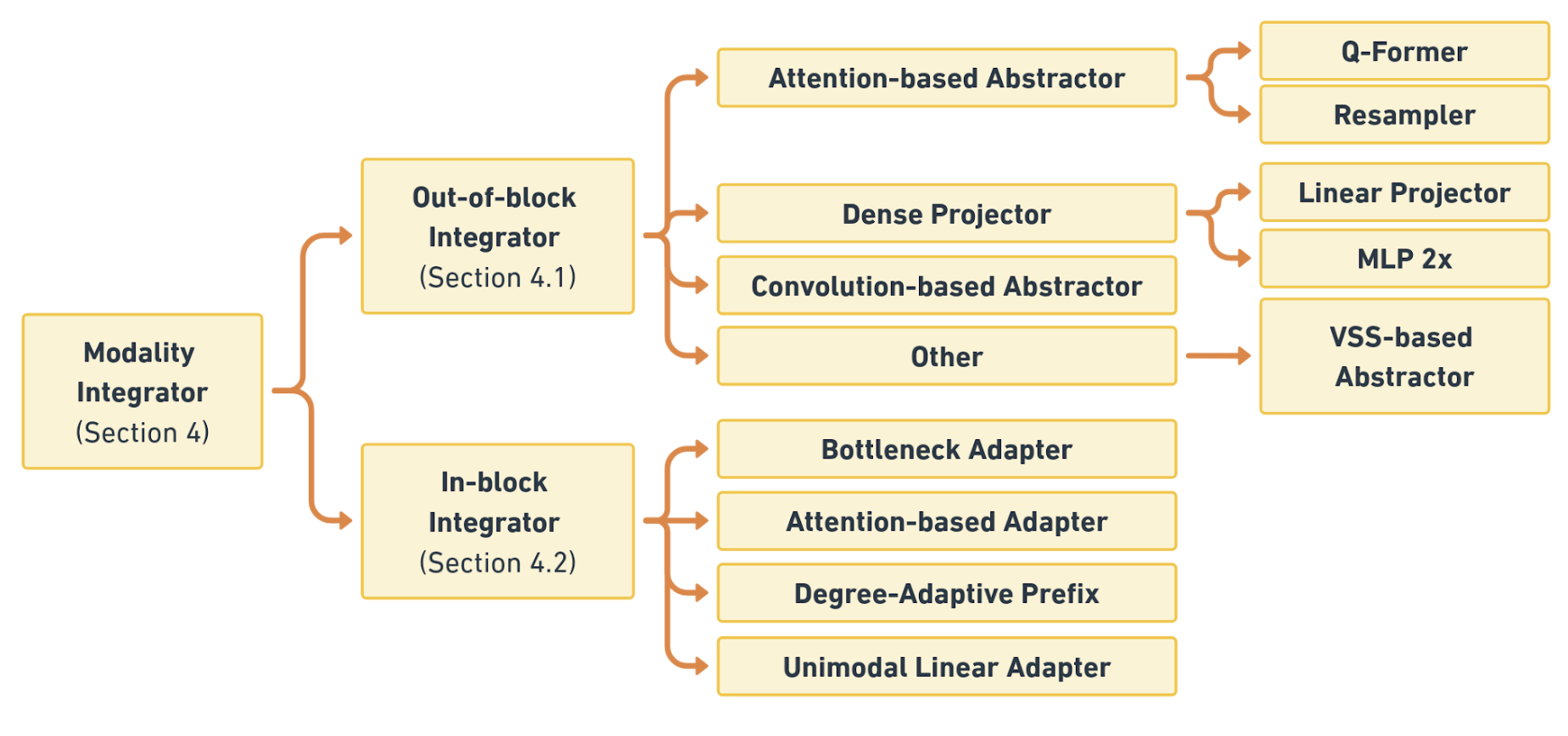}
    \caption{\textbf{Taxonomy of Modality Integrator.}}
    \label{fig:MI_Taxonomy3}
\end{figure}

The modality integrator is categorized into Out-of-block Integrators and In-block Integrators, where the "block" refers to the LLM. The Out-of-block Integrators, as a basic component of VLLM, align the visual features extracted by vision encoders with the input of LLMs. The In-block Integrators are modules inserted into the LLM architecture, which change the computational graph of LLMs and fuse the multimodal information. The structure of the Modality Integrator is crucial for efficiently integrating the vision modality into the LLM, as it directly impacts the model's ability to process and understand multimodal information and the trainable parameter scale. The taxonomy of MI is shown in Figure \ref{fig:MI_Taxonomy3}.

\subsection{Out-of-block Integrator}
\label{sec: Out-of-block Integrator}
Commonly, the Out-of-block Integrator is the external connector between the vision encoder and LLM, transforming the visual features over length or dimension. Based on the architecture, the Out-of-block Integrator is classified into an Attention-based Abstractor, Dense Projector, Convolution-based Abstractor, and VSS-based Abstractor.

\begin{table*}[!htb]
\small
\centering

    \begin{tabular}{lccc}
        \toprule
        & \textbf{Attention-based} & \textbf{Linear Mapping} & \textbf{Convolution-based} \\
        \midrule
        Maintaining Local Context & & \checkmark & \checkmark \\
        Maintaining Global Context & \checkmark & & \\
        Adaptability & \checkmark & & \checkmark \\
        Parameter efficiency & & \checkmark & \checkmark \\
        \bottomrule
    \end{tabular}
    
     \caption{\textbf{Comparison of Out-of-block Integrators.} Adaptability refers to the flexibility in adjusting the output token length. The parameter efficiency is compared under the same number of layers.}
    \label{tab: connector comparison} 
\end{table*}

\subsubsection{Attention-based Abstractor}
\label{sec: Attention-based Integrator}
\paragraph{Resampler}
\label{sec: Resampler}
The Resampler bridges the vision encoder and LLM by inputting visual embedding $X_v$ and fixed-length learnable prompts $P_v'$ into $L$ layers self-attention blocks and outputting the $P_v$ carrying visual information.

In the first transformer block of the Resampler:
\begin{equation}
\begin{aligned}
    h_1&=\text{Block}_1([X_v;P_v']),\\
    Q&=P_v'W^Q, K=[X_v; P_v']W^K,V=[X_v; P_v']W^V ,\\
    h_1'&=\text{Attention}(Q,K,V)+P_v',\\
    h_1&=  h_{1}'W_\theta^{FFN}+  h_{1}'.
\end{aligned}
\end{equation}

In the last transformer layer,
\begin{equation}
    P_v=\text{Block}_{L_i}(h_{L_i-1}),
\end{equation}
where   $ P_v' \in \mathbb{R}^{p \times d_v},P_v\in \mathbb{R}^{p \times d_t} $.  

ClipCap \cite{mokady2021clipcap} and Flamingo \cite{alayrac2022flamingo} utilize the Resampler as a prefix former, meaning that $P_v$ is fed into each transformer block of the language model. Whereas Meta-Mapper \cite{najdenkoska2023meta} and Qwen-VL \cite{bai2023qwen} regard it as a soft prompt former, $P_v$ is prepended to the textual input for the language generator. In Meta-Mapper, the resampler is self-attended, where the input of $Q,K,V$ are all $[X_v;P_v]$.  mPLUG-Owl2 forms the Resampler as a self-attention layer with a SwiGLU activation function \cite{shazeer2020glu}.

There are two advantages of the Resampler. First, it shortens the length of visual input to the language model and keeps the input length constant, irrespective of the variable length of the original visual embedding \cite{alayrac2022flamingo}. This enables flexibility in the number of images, which is significant in video input. Second, the transformer architecture is more expressive than the linear projection \cite{mokady2021clipcap}, which can capture more representative visual information. 

However, Cha et al. \cite{cha2024honeybee} argue that self-attention-based Resampler loses visual information because it puts more attention weight on the major object of an image while ignoring insignificant objects. To strengthen the spatial context-capturing capability of the Attention-based Abstractor, Deformable attention \cite{zhu2020deformable} is utilized in the D-Abstractor, considering the reference point and learnable offset. 

Another drawback is the parameter inefficiency of attention-based structures. To save parameters, MAPL \cite{manas2022mapl} adds down projection layers before the Resampler to reduce the input dimension of the visual embedding $X_v$ and up-project the output $P_v$ to expand its dimension, thus causing a considerable reduction in parameter consumption. The process can be formulated as:
\begin{equation}
\begin{aligned}
    X_{\text{input}} &= [X_v W_{\text{down}}; P_v], \quad X_{\text{output}} = P_v W_{\text{up}} \\
    W_{\text{down}} &\in \mathbb{R}^{d_v \times r}, \quad W_{\text{up}} \in \mathbb{R}^{r \times d_v}, \quad \text{where } r < d_v.
\end{aligned}
\end{equation}

\paragraph{Q-Former}
Unlike the Resampler, the Q-former proposed by BLIP-2 \cite{li2023blip} involves visual and textual input. Each layer of the Q-former is composed of a self-attention module shared across the learned queries $P_{tv}$ and the input text $X_t$, a cross-attention layer between image embedding $X_v$ and the soft prompt $P_{tv}$, and two feed-forward layers for two modalities separately. The feed-forward process is formulated as below.

In the first layer of the Q-former,
\begin{equation}
     h_1=\text{Block}_1([X_v;P_{tv}';X_t]).
\end{equation}

For each layer in Q-former, there is a shared self-attention layer between text $X_t$ and learned query $P_{vt}$, where
\begin{equation}
\begin{aligned}
    &Q=[P_{tv}';X_t]W^Q, K=[P_{tv}';X_t]W^K, V=[P_{tv}';X_t]W^V\\
    &h_{1t}'= \text{Attention}(Q,K,V),
\end{aligned}
\end{equation}
After the shared self-attention module, the two modalities are separated and processed. For the vision stream, the soft prompt is cross-attended with visual embedding $X_v$ and processed by the visual feed-forward layer. The $h_{tv} $ means that the output of this block contains both textual and visual information.

\begin{equation}
\begin{aligned}
    &Q'=P_{tv}'W^Q, K'=X_vW^K, V' = X_vW^V,\\
    &h_{1tv}' = \text{Attention}(h_{1t}' ,X_v),\\
    &h_{1tv} = h_{1t}' W_{\theta}^{FFN_v}.
\end{aligned}
\end{equation}

For the text stream, the self-attended $h_{1t}'$ is processed by the textual feed-forward layer.

\begin{equation}
h_{1t} = h_{1t}' W_{\theta}^{FFN_t}.
\end{equation}

In the last layer,
\begin{equation}
\begin{aligned}
    &P_{tv}=\text{Block}_{L}(h_{L-1tv}),\\
    &h_t=\text{Block}_{L}(h_{L-1t}).
\end{aligned}
\end{equation}

In the first pretraining stage, the Q-former is pre-trained by three tasks to fuse the two modalities: Image text matching, Image contrastive learning, and Image grounded text generation, to align the visual soft prompt to the LLM's encoding space.

In the second pretraining stage, a dense layer is trained to project to the language model input dimension. The parameters are updated by language modelling loss.
\begin{equation}
    P_{tv} =\text{Q-Former} (X_v) W_\theta,
\end{equation}
where $ X_v \in \mathbb{R}^{n_v \times d_v}, P_{tv}\in \mathbb{R}^{n_p  \times d_t}.$
BLIP2 pretraining only involves the Q-former and a dense layer, thus saving over 96\% of the parameters.

Frozen Q-former is further adopted in Mini-GPT4 \cite{zhu2023minigpt} and X-LLM \cite{chen2023x} to extract text-conditioned visual embedding. X-LLM and Video-Llama \cite{zhang2023video} also use a trainable Q-Former with a linear layer as the integrator to augment visual and audio perception.

\subsubsection{Dense Projector}
\label{sec: Dense Integrator}
 A Dense Projector is either a single linear layer or a two-layer MLP, which can be formulated as:
\begin{equation}
    P_v=X_v   W ,
\end{equation}
\begin{equation}
    P_v =  \sigma( X_v W_1 ) W_2  .
\end{equation}

It has been proved that a trainable linear projector is capable of translating visual semantic information to an LM-understandable language and can gain comparable performance with end-to-end trained VLMs \cite{merullo2022linearly}.  FROMAGe \cite{koh2023grounding} uses linear layers to form visual soft prompts for the image captioning tasks and form both textual and visual soft prompts for the ITM tasks. LLaVA \cite{liu2024visual} uses an MLP to connect frozen LLM and vision encoder. MiniGPT-4 \cite{zhu2023minigpt} adds a trainable linear layer to connect the frozen vision encoder, Q-former, and LLM, achieving high parameter efficiency. 

However, Dense Projector lacks the flexibility to adjust the length of visual representation. To overcome this, the projector output in MemVP \cite{jie2024memory} is directly fed into the FFN of the LLM transformer block without occupying the input tokens.

In addition to the standalone use of Dense Projectors, its combination with In-block Integrators is frequently employed to achieve deeper modality interaction \cite{sung2022vl,wang2023cogvlm,zhang2023llama,eichenberg2021magma,yang2022zero,hu2023vl}.

\subsubsection{Convolution-based Abstractor}
\label{sec: Convolution-based Integrator}
Cha et al. \cite{cha2024honeybee} spots the limitation of the current Dense Projectors and Attention-based Abstractors. The linear mapper is good at retaining complete visual context while having no flexibility in controlling the output number of visual tokens, which affects the inference efficiency of the MLLM. The attention-based integrator can control the visual token number, while the attention mechanism extracts only the representative subject in the image, leading to visual information loss. To retain both properties, it proposes a C-Abstractor, the architecture of which includes two modules $H$ and $Q$, which are both composed of $N $ ResNet blocks \cite{xie2017aggregated}. They are also connected by an adaptive average pooling layer. The architecture can be formulated as:
\begin{equation}
\begin{aligned}
  H_1&=\text{ResBlock}_1({X_v}) = \text{ReLU}({W}_{\text{BatchNorm}}({X_v} \cdot {W}_{\text{Conv}})),\\
     H_N&=\text{ResBlock}_L({H_{N-1}}),\\
     P &=\text{AdaptiveAvgPool}({H_{N}}), \\
    Q_1&=\text{ResBlock}_{N+1}(P)  = \text{ReLU}({W}_{\text{BatchNorm}}(P \cdot {W}_{\text{Conv}})),\\
      Q_N&=\text{ResBlock}_{2N}(Q_{N-1}) .\\
\end{aligned}    
\end{equation}

MobileVLM series \cite{chu2023mobilevlm,chu2024mobilevlm} efficiently project the visual embedding $X_v$ through depthwise-convolution $ {W}_{\text{DepthWise}} $ and average pooling $\text{AvgPool}_{2 \times 2}$.
For the MobileVLM v1, the Out-of-block Integrator can be formulated as:
\begin{equation}
\begin{aligned}
    \mathbf{P}_{v}  = 
    \begin{cases}
        P_{v}'' &= {W}_{\text{PatchWise}}(\text{GELU}({W}_{\text{PatchWise}}(X_v))), \\
        P_v' &= {W}_{\text{LayerNorm}}[{W}_{\text{DepthWise}}({W}_{\text{LayerNorm}}({W}_{\text{DepthWise}}(P_{v}'')))] + P_{v}'', \\
        {P}_{v} &= {W}_{\text{LayerNorm}}[{W}_{\text{DepthWise}}({W}_{\text{LayerNorm}}({W}_{\text{DepthWise}}( P_v')))],
    \end{cases}
\end{aligned}
\end{equation}

where $GELU$ refers to the non-linear activation function, the input and output sizes are $ X_v \in \mathbb{R}^{n_v \times d_v}, P_{v}\in \mathbb{R}^{(n_v/4)  \times d_t} $.

MobileVLM v2 adds a Positional Encoding Generator \cite{chu2021conditional} and a residual connection, enabling the architecture to replace one Depth Convolutional layer to average pooling, which reduces over 99\% parameters compared to the v1 projector. 
\begin{equation}
    \begin{aligned}
    \mathbf{P}_{v} = 
    \begin{cases}
        P_{v}'' &= {W}_{\text{PatchWise}}(\text{GELU}({W}_{\text{PatchWise}}(X_v))), \\
        P_v'&= \text{AvgPool}_{2\times2}(P_{v}''), \\
        \mathbf{P}_{v} &= {W}_{\text{DepthWise}}(P_v') + P_v'.
    \end{cases}
\end{aligned}
\end{equation}

The input and output sizes are $ X_v \in \mathbb{R}^{n_v \times d_v}, P_{v}\in \mathbb{R}^{n_v/k^2  \times d_t} $, where $k$ stands for the average kernel size.

\subsubsection{Other}
\label{sec: vss-based}
VL-Mamba \cite{qiao2024vl} proposes a Vision Selective Scan (VSS) mechanism to capture richer information from the non-causal visual data without increasing parameters.  It concatenates the feedforward, backward, horizontal, and vertical scan of image patches and uses an MLP layer or linear layer to process them.

\subsection{In-block Integrator}
\label{sec: In-block Integrator}
The In-block Integrator here refers to tunable modules inserted into transformer blocks of LLMs. By changing the computational graph of the LLM, the In-block Integrator further fuses the two modalities or controls the degree of the introduced visual information. Commonly, In-block Integrators are adopted together with the external connector. Based on the architecture, the In-block Integrator is classified into Degree-Adaptive Prefix, Bottleneck Adapter, Attention-based Adapter, and Unimodal Linear Adapter.

\subsubsection{Bottleneck Adapter}

As described in \ref{sec: Bottleneck Adapter}, the Bottleneck Adapter \cite{houlsby2019parameter} is a typical structure for parameter efficiency. Sung et al. \cite{sung2022vl} find that in the multimodal context, the adapter carries the information of the introduced modality instead of task-specific knowledge.

MAGMA \cite{eichenberg2021magma} first attempts to utilize both a Dense Projector and a Bottleneck Adapter. VL-PET \cite{hu2023vl} proposes four adapter architectures of three parameter sizes. In addition to typical bottleneck architecture, it also adopts a down-projection layer ${W}_{\text{down}}' \in \mathbb{R}^{d \times 1}
$ and copies projected embeddings across the dimension $N$ times to expand to ${W}_{\text{down}} \in \mathbb{R}^{N \times d}$. In this way, only the parameter $W_\text{down}'$ is tunable. To deal with mixed modality input, LaVIN \cite{luo2024cheap} proposes a modality classifier to shift between single-modal and multimodal processing adapters automatically. To reduce the parameter, the two adapters share down-sampling projector weights during finetuning.

\subsubsection{Attention-based Adapter}
\label{sec: Attention-based Adapter}
The base structure of the Attention-based Adapter is the transformer block as described in \ref{sec: transformer}.
In Flamingo \cite{alayrac2022flamingo}, a fusion adapter called Gated XAtten-Dense is inserted before each LLM transformer block, fusing the visual soft prompt formed by Resampler and the text embedding. The adapter cross-attends two modalities by retrieving text-conditioned visual information. It is composed of a cross-attention sublayer and a feed-forward layer, with TANH gating after each sublayer. For the first adapter layer, the process can be formulated as:
\begin{equation}
\begin{aligned}
      Q &=  X_t  W^Q,\quad K = P_v  W^K,\quad  V = P_v  W^V, \\
       h_i'&=\text{TANH}(\text{Attention}(Q,K,V))+X_t,\\
         h_i&=\text{TANH}(h_i'W_\text{FFN})+h_i',
\end{aligned}
\end{equation}

where $\text{TANH}(x) = (e^x - e^{-x}) / (e^x + e^{-x})$.

QaP \cite{liang2024querying} inserts Resampler (Section \ref{sec: Resampler}) inside the LLM, between the frozen self-attention module and FFN, processing the constant modality query $P'$. Unlike the common practice, QaP doesn't involve an Out-of-block Integrator but utilizes a standalone internal Resampler to process prefix. It achieves higher parameter efficiency by setting the learnable prefix length as one for each modality. 

Considering parameter efficiency and memory efficiency, LST \cite{sung2022lst} tunes a side network composed of the down-scaled LLM blocks. The weights of each transformer block are shortened proportionally at the dimension level. In this way, not only are the tuned parameters reduced, but the weights are back-propagated through the side network, thus saving the memory for LLMs.

\subsubsection{Degree-Adaptive Prefix}
\label{sec: degree adaptive prefix}
Unlike other VLLMs where the prefix is transformed from visual embeddings by Out-of-block Integrators, Degree-Adaptive Prefix is internally informed by LLMs, with their degree controlled by a gating factor \cite{zhang2023llama, gao2023llama, liang2024querying}. In QaP, the prefix for added modality output by $i$-th Resampler can be formulated as: 
\begin{equation}
\begin{aligned}
    &Q =  P'  W^V ,\quad K = X_v   W^K,\quad  V = X_v  W^V,\\
    &P_i=g_i(\text{Resampler}(Q,K,V))+X_v,
\end{aligned}
\end{equation}

where $g_i $ represents a tunable gating parameter controlling the degree of modality information.

Considering the disturbance a randomly initialized prefix may cause to the LLM, LLaMA-Adapter \cite{zhang2023llama} proposes a gating parameter to control the weight of the learnable prefix contributing to the next token prediction. Prefix is added to each of the top-$l$ transformer layers, and its disturbance can be minimized by the zero-initialized gating factor at the early training stage. To compute the effect of the prefix token, at each transformer layer, the hidden states of the predictive word are generated, and the effect is quantified by the attention score before softmax. The visual soft prompt is formed by a trainable linear projection and element-wisely added to the prefix at each transformer layer. Suppose the current timestep is $k$, the attention score of  $l$-th transformer layer can be formulated as:
\begin{equation}
\begin{aligned}
    Q_l& =  X_{k}^l W_\phi^Q,\quad K_l = [P_v^l,X_{<k}^l;X_{k}^l] W_\phi^K,\quad  V_l = [P_v^l;X_{<k}^l;X_{k}^l]  W_\phi^V, \\
    S_l&(P_v^l,X_{<k}^l,X_{k}^l, W_\phi) = QK^T / \sqrt{D} \in \mathbb{R}^{1 \times (m+n+1)},\\ 
    S_l& = [S_l^m ; S_l^{n+1}]^T,\\
    S_l^q& = [\text{softmax}(S_l^m) \cdot g_l ; \text{softmax}(S_l^{n+1})]^T.
\end{aligned}
\end{equation}

The trainable parameters are the prefix $P_v$, gating parameters $g$, and the linear projection $W_\theta$.

\subsubsection{Unimodal Linear Adapter}
\label{sec: Unimodal Linear Adapter}
Unimodal Linear Adapters are defined as the additional sets of linear layers used to form query, key, and value for the added modality. Compared to mapping the visual features into discrete textual representation space, this method avoids ignoring the rich semantic information in the visual \cite{ye2024mplug}.
COGVLM \cite{wang2023cogvlm} adds a set of query, key, and value projection Layers and a trainable FFN block to process the visual modality. The process can be formulated as: 

\begin{equation}
\begin{aligned}
    &Q = [P_v  W_{v}^Q; X_t  W_t^Q],\quad K = [P_v  W_v^K; X_t W_t^K],\quad  V = [P_v  W_v^V; X_t  W_t^V],\\
    &\text{FFN}(S_v,S_t) = [\text{FFN}_v(S_v); \text{FFN}_t(S_t)].
\end{aligned}   
\end{equation}
mPLUG-Owl 2\cite{ye2024mplug} adopts a similar idea in its Modality-Adaptive Module, and the difference is that no $W_v^Q$ is added.

\section{ Training Paradigms, Datasets, and Experiment}
\label{sec: training paradigm}
In this section, we first introduce evaluation benchmarks applicable across all three training paradigms, providing a consistent framework for comparison. Next, we delve into the three primary training paradigms: Single-stage Tuning, Two-stage Tuning, and Direct Adaptation, as illustrated in Figure \ref{fig: Training Paradigm}. For each paradigm, we discuss the corresponding training techniques, datasets, parameter-efficient strategies, and performance discussion. Notably, the experiments for VLLMs following the Direct Adaptation paradigm are replicated, and those for other paradigms are sourced from their respective publications or GitHub repositories.

\begin{table}[!htb]
\fontsize{12pt}{12pt}\selectfont
    \centering
    \renewcommand{\arraystretch}{1.2} % 行高调整
\setlength{\tabcolsep}{4pt}       % 列间距调整

\resizebox{\textwidth}{!}{
    \begin{NiceTabular}{p{2.5cm}lp{2cm}p{3.5cm}p{2cm}p{2.3cm}p{2cm}lllp{3cm}l}
    \toprule
    \multirow{2}{*}{\textbf{Dataset}} & \multicolumn{3}{c}{\textbf{Format}}         & \multicolumn{2}{c}{\textbf{Size}}  & \multicolumn{1}{c}{\textbf{Evaluation}} & \multicolumn{3}{c}{\textbf{Capability}}   & \multicolumn{2}{c}{\textbf{Basic Info}} \\
 &\textbf{ Task} & \textbf{Input} & \textbf{Label} & \textbf{Image \#} & \textbf{Question \# }& \textbf{Matrix} & \textbf{Perception} & \textbf{Reasoning} & \textbf{OCR} &  \textbf{Source} &  \textbf{Time} \\ \midrule
\textbf{Traditional} & ~ & ~ & ~ & ~ & ~ & ~ & ~ & ~ & ~ & ~ & ~ \\ 
\underline{VQAv2} \cite{2015VQAv2} & VQA (short) & $[I,Q]$ & Phrases  & 82783/ 40504/ 81434 & 443757/ 214354/ 447793 & acc. & Basic & - & ~ & COCO & 2017 \\
OKVQA \cite{okvqa2019} & VQA (short) & $[I,Q] $ & Phrases & - & 9009/5046 & acc. & Advanced & Advanced & ~ & COCO & 2019 \\ 
\underline{GQA} \cite{hudson2019gqa} & VQA (short) & $[I,Q] $ & Phrases & 113018 & 22000000 & acc. & Basic & Basic & ~ & Visual Genome & 2019 \\ 
VisWiz \cite{Gurari2018VizWizGC} & VQA (short) & $[I,Q] $ & Phrases & 20523/ 4319/ 8000 & 20523/ 4319/ 8000 & acc. & Advanced & Basic & \tick & Photo taken by blind people & 2023 \\ 
TextVQA \cite{TextVQA2019} & TextVQA (short) & $[I,Q] $ & Phrases  & 28408 & 45336 & acc. & Basic & - & \tick & Open Images v3 \cite{ krasin2017openimages} & 2024 \\
OCRVQA \cite{OCRVQA2019} & TextVQA (short) & $[I,Q] $ & Phrases  & 207572 & 800000/ 100000/ 100000 & acc. & Basic & Basic & \tick & \cite{ iwana2016judging} & 2024 \\ 
\underline{ScienceQA} \cite{lu2022learn} & VQA(MC) & $[I,Q,C,M]$ & MC  & - & 21208 & acc. & Basic & Advanced & ~ & Scientific problems & 2022 \\ 
\underline{COCO} Captions \cite{chen2015microsoft} & ImageCap & $[I]$ & Free-Form Captions & 82783/ 40504/ 40775 &  413915/ 202520/ 379249 & CIDEr \cite{vedantam2015cider}/ BLEU@4 \cite{papineni2002bleu} & Basic & - & ~ & COCO & 2015 \\ 
Nocaps \cite{agrawal2019nocaps} & ImageCap & $[I]$ & Free-Form Captions & / 4500/ 10600 & 166100 & CIDEr, SPICE \cite{anderson2016spice}   & Basic & - & ~ & COCO, Open Images & 2019 \\
Flickr30K \cite{Flicker30k2015} & ImageCap & $[I]$ & Free-Form Captions & 29783/  1000/  1000 & 148915/ 5000/ 5000 & CIDEr/ BLEU@4 & Basic & - & ~ & ~ & 2015 \\ 
\underline{NLVR} \cite{suhr2017NLVR}  & NLVR & $[I_1,I_2,Cap]$ & Binary & ~ & 74460/ 5940/ 11844 & acc. & Basic & Basic & ~ & Google Images & 2019 \\ \midrule

\textbf{Advanced} & ~ & ~ & ~ & ~ & ~ & ~ & ~ & ~ & ~ & ~ & ~ \\ 
\underline{MMBench} \cite{liu2023mmbench} & VQA(MC) & $[I,Q,M]$ & MC  & - & 3217 & CircularEval Top-1 acc. & Advanced & Advanced & \tick & Various sources & 2024 \\ 
\underline{SEED-I} \cite{li2023seed} & VQA(MC) & $[I,Q,M]$ & MC  $^\dagger$ & - & 15973 & acc. & Basic & Advanced & \tick & CC3M & 2023 \\ 
LLaVA-Bench \cite{liu2024visual} & VQA (open) & $[I,Q]$ & Free-Form Response $^\dagger$ & 30 & 90 & GPT-4 score & Advanced & Advanced & ~ & COCO & 2023 \\ 
LLaVA-Bench (In-the-Wild) \cite{liu2024visual} & VQA (open) & $[I,Q]  $ & Free-Form Response $^\dagger$ & 24 & 60 & GPT-4 score & Advanced & Advanced & ~ & - & 2023 \\ 
OwlEval \cite{ye2023mplug} & VQA (open) & $ [I,Q] $ & Free-Form Response & 50 & 82 & Manual evaluation  \cite{wang2023selfinstruct} & Advanced & Advanced &  \tick & MiniGPT-4, MM-REACT, BLIP-2, GPT-4 & 2023 \\ 
MM-Vet \cite{yu2024mm} & VQA (open) & $[I,Q]$ & Phrases/ Free-Form Response & 200 & 218 & GPT-4 Score & Advanced & Advanced & \tick & Online sources, VCR \cite{2019VCR}, ChestX-ray14 \cite{wang2017chestx} & 2023 \\ 
\underline{MME} \cite{fu2023mme} & VQA (Binary) & $[I,Q]$ & Binary  & 1077 & 2194 & acc. & Advanced & Advanced & \tick & COCO & 2024 \\ 
POPE \cite{Lihallucination2023} & VQA (Binary) & $[I,Q] $ & Binary  & 40504 & - & F1 score & Basic & - &  & COCO & 2024 \\ 
LLVisionQA  \cite{wu2024qbench} & VQA (MC) & $ [I,Q,M]$ & MC  & 2990 & 2990 & GPT score & - & - &  & 10 datasets & 2023 \\ 
LLDescribe  \cite{wu2024qbench} & VQA (open) & $[I,Q]$ & Free-Form Captions & 499 & 499 & GPT-4 Score & - & - &   & 10 datasets & 2023 \\ 
QBench-Assessment  \cite{wu2024qbench} & VQA (short) & $[I,Q]$ & Number & 81,284 & - & Softmax & - & - &  & 7 datasets & 2023 \\ \bottomrule
    
    \end{NiceTabular}
    }
\caption{\textbf{The Summary of Benchmarks.} In the Input column, $I$ denotes image, $Q$ denotes question, $C$ denotes the context, and $M$ denotes the MC options.  $^\dagger$ denotes GPT-4 Is involved in label generation. Otherwise, the label is created manually. The Image \# and Question \# show train, validation, and test split statistics if available. The benchmarks used for performance analysis are underlined. }
\label{tab: Evaluation Data}
\end{table}

\subsection{Evaluation Benchmarks}

The benchmarks are summarized in Table \ref{tab: Evaluation Data}. These benchmarks test two main visual capabilities: Perception ability and Reasoning ability. Both abilities are further categorized into basic and advanced, depending on whether they require knowledge beyond the information available from the image itself. Perception is the capability to recognize features and information from visual input. The MLLMs with basic perception capability can capture features such as color, count, and position \cite{2015VQAv2}, while those with advanced capability can utilize the knowledge in LLM to recognize image emotion \cite{liu2023mmbench}, movie posters, celebrities \cite{fu2023mme} and so on. Reasoning ability is the capability to make correct judgments through logical reasoning, object relationship understanding, and attribute inference when faced with complex problems. Testing basic reasoning ability involves understanding the complex question \cite{hudson2019gqa}, while the advanced involves utilizing common sense and prior knowledge in LLM \cite{lu2022learn,li2023seed}.

Apart from visual understanding, language generation capability can be evaluated through tasks requiring free-form output, such as visual question answering (VQA) with open-ended responses and image captioning. While traditional benchmarks evaluate this capability using statistical similarity metrics such as CIDEr and BLEU \cite{vedantam2015cider,papineni2002bleu}, advanced benchmarks have shifted towards assessing richer and more detailed outputs \cite{liu2024visual,ye2023mplug,yu2024mm}. For example, some advanced benchmarks incorporate evaluations conducted by GPT-4 \cite{liu2024visual,yu2024mm,wu2024qbench}, offering a qualitative, context-sensitive measure of the models’ ability to produce detailed and accurate content.

Traditional benchmarks are applicable to all three training paradigms, offering a consistent foundation for evaluation. In contrast, advanced benchmarks are widely used in instruction-tuned models within the Single-stage Tuning paradigm \cite{zhang2023llama,liu2024visual, zhu2023minigpt,ye2023mplug,gao2023llama,zhang2023video,bai2023qwen,liu2024improved,wang2023cogvlm,ye2024mplug,cha2024honeybee,chu2023mobilevlm,chu2024mobilevlm,qiao2024vl}. This is because these models can follow instructions and align outputs closely with the tasks defined in the benchmarks.

Various benchmarks have proposed distinct methods to enable evaluation matrices to assess model outputs. For tasks requiring phrase outputs, accuracies are averaged across all combinations of human annotator subsets in VQAv2 \cite{2015VQAv2}. For multiple-choice (MC) tasks, varied post-processing strategies are employed to align outputs with task requirements. For instance, MemVP \cite{jie2024memory} limits the output length to one for MC tasks, while LLaVA \cite{liu2024visual} introduces instructions such as: "Answer the question using a single word or phrase" or "Answer with the option's letter from the given choices directly." However, recent studies \cite{ye2023mplug,wu2024qbench} indicate that few VLLMs consistently generate outputs in the instructed format. As a result, there is a growing trend to incorporate LLMs into the evaluation process. For instance, MM-Bench \cite{liu2023mmbench} utilizes both rule-based and LLM-driven choice extraction methods to identify answers from the generated textual outputs. Additionally, Q-bench \cite{wu2024qbench} develops a GPT-assisted evaluation framework comprising five iterative rounds.

\begin{table}[!htb]
\fontsize{13pt}{15}\selectfont

    \centering
    \renewcommand{\arraystretch}{1.2} % 行高调整
\setlength{\tabcolsep}{4pt}       % 列间距调整

    \resizebox{\textwidth}{!}{
    \begin{NiceTabular}{llp{5cm}p{5cm}llcccccc}
    \toprule
        \textbf{Model} & \textbf{LLM} & \textbf{Trainable} & \textbf{Finetune Technique} & \textbf{Para \#} & \textbf{Para \%} & \textbf{VQAv2} & \textbf{OKVQA} & \textbf{GQA} & \textbf{COCO} & \textbf{SQA-IMG} & \textbf{TextVQA} \\ \midrule
        \textbf{Single Stage Training} & ~ & ~ & ~ & ~ & ~ & ~ & ~ & ~ & ~ & ~ & ~ \\ 
        \underline{Frozen} \cite{tsimpoukelli2021multimodal} & GPT2-like  1.5B& Linear & - & 40.3M & 0.64\% & 48.4 & 19.6 & - & - & - & - \\ 
        ClipCap \cite{mokady2021clipcap} & GPT-2 1.5B & Resampler & - & 43M & - & - & - & - & 113.08 & - & - \\ 
        MAGMA \cite{eichenberg2021magma} & GPT-J 6B& Linear Projector + Bottleneck Adapter  & - & - & - & \cellcolor{blue!15}61.5 & 40.3 & \cellcolor{blue!15}49.6 & 57 & - & - \\ 
        \underline{Flamingo} \cite{alayrac2022flamingo} &  Chinchilla & Resampler + Attention-based Adapter  & - & - & - & - & \cellcolor{blue!15} 44.7 & - & 79.4 & - & \cellcolor{blue!15}31.8 \\ 
        MAPL \cite{manas2022mapl} & GPT-J 6B & Resampler & - & 3.4M & 0.05\% & 43.51 & 18.27 & - & \cellcolor{blue!15}125.2 & - & 10.99 \\ 
        ESPER \cite{ESPER2023Yu} &GPT-2 117M &MLP 2x &-&8M &2.6\% &- &- &- &78.2 &- &-\\
        BLIP-2 \cite{li2023blip} & OPT 6.7B  & Q-Former + Linear & - & 108M & 1.38\% & 52.6 & 36.4 & 36.4 & - & -& - \\ 
        ~ & ~ & ~ & ~ & ~ & ~ & \underline{51.5} & \underline{31.9} & \underline{43} & \underline{93.67} & ~ & \underline{21.3} \\ \midrule
       
        \textbf{Two-stage Training} & ~ & ~ & ~ & ~ & ~ & ~ & ~ & ~ & ~ & ~ & ~ \\ 
        \underline{Flamingo} \cite{alayrac2022flamingo} &  Chinchilla & Resampler + Attention-based Adapter  & Per-task & - & - & 82.1 & - & - & 138.1 & - & 54.1 \\ 
        LLaMA-Adapter \cite{zhang2023llama} & LLaMA  & Linear + Adaptive Prefix & Instruction Tuning (T+V) & 1.8M & - & - & - & - & - & 80.32 & - \\ 
        ESPER \cite{ESPER2023Yu} &GPT-2 117M &MLP 2x &Per-task &8M &2.6\% &- &- &- &103.1 &- &-\\
        LLaVA \cite{liu2024visual} & LLaMA  & Linear + LoRA & Instruction Tuning (V) & 4.4M & -& - & - & - & - & \cellcolor{pink!30}90.28 & - \\
        MiniGPT4 \cite{zhu2023minigpt} & Vicuna & Linear & Instruction Tuning (V) & - & - & - & - & 32.2 & - & - & - \\ 
        BLIP-2 \cite{li2023blip} & OPT6.7B & Q-Former + Linear & Per-task & 1.1B & 14\% & \cellcolor{pink!30}82.19 & - & - & \cellcolor{blue!15}\textbf{145.2} & - & - \\ 
       LLava 1.5 \cite{liu2024improved} & Vicuna v1.5  & MLP+LoRA & Instruction Tuning (V) & 0.3B & 4.61\% & 79.1 & - & \cellcolor{blue!15}\textbf{63} & - & 68.4 & \cellcolor{pink!30}58.2 \\
        ~ & ~ & MLP+DoRA & ~ & 0.3B & 4.63\% & 78.6 & ~ & \cellcolor{pink!30}62.9 & - & - & - \\ 
        CogVLM-Chat \cite{wang2023cogvlm} &  Vicuna-v1.5 & MI+VE & Multi-task Learning  + Instruction Tuning (V) & - & - & \cellcolor{blue!15}\textbf{82.3} & \cellcolor{blue!15}\textbf{64.8}& - & - & \cellcolor{blue!15}91.2 & \cellcolor{blue!15}\textbf{70.4} \\ 
        MobileVLM \cite{chu2023mobilevlm} & MobileLLaMA 2.7B & Convolution + Small LLM & instruction tuning (V) & 2.71B & - & - & - & 59 &- & 61 & 47.5 \\ 
        ~ & ~ & Convolution + LoRA & ~ & 0.2B & 7.41\% & - & - & 58.4 & - & 59.0 & 46.7 \\ 
        MobileVLM v2 \cite{chu2024mobilevlm} & MobileLLaMA 2.7B  & Convolution + Small LLM & Multi-task Learning  + Instruction Tuning (V) & 2.7B & - & - & ~ & 61.1 & - & 70.0 & 57.5 \\ 
        \underline{VL-Mamba} \cite{qiao2024vl} & Mamba LLM-2.8B & VSS +Small LLM & Instruction Tuning (V) & - & - & 76.6 & - & 56.2 & - & 65.4 & 48.9 \\ 
        ~ & ~ & ~ & ~ & ~ & ~ & \underline{80.1} & \underline{64.8} & \underline{56.1} & \underline{141.65} & \underline{76.1} & \underline{55.82} \\ \midrule
        \textbf{Direct Adaptation} & ~ & ~ & ~ & ~ & ~ & ~ & ~ & ~ & ~ & ~ & ~ \\ 
        VL-Adapter \cite{sung2022vl} & T5 base 220M & Linear + Bottleneck Adapter & Multi-task Learning & - & 7.98\% & 66.99 $^\dagger$  & - & \cellcolor{pink!30}56.36 $^\dagger$  & 111.85 $^\dagger$  & - & - \\ 
        \underline{PromptFusion} \cite{liang2022modular} & GPT3 175B & Prompt & Per-task & 15K & - & 34.1 & -& - & - & - & - \\
        LST \cite{sung2022lst}  & T5 base 220M & Linear + Attention-based Adapter & Multi-task Learning & - & 7.46\% & \cellcolor{blue!15}67.22$^\dagger$  & ~ & \cellcolor{blue!15}56.37$^\dagger$  & 115.05$^\dagger$  & - & - \\ 
        eP-ALM \cite{shukor2023ep} & OPT 2.7B & Linear & Per-task & 4.2M & - & 54.89 & - & 42.91 & 111.63 & - & - \\ 
        LaVIN \cite{luo2024cheap}    &  LLaMA & Linear + Bottleneck Adapter & Instruction Tuning (T+V) & 3.8M & - & - & - & - & - & 74.52$^\dagger$ & - \\ 
        VL-PET \cite{hu2023vl}  & T5 base 220M  & Linear + Bottleneck Adapter & Multi-task Learning & - & 7.31\% & \cellcolor{pink!30}67$^\dagger$ & - & 55.97$^\dagger$ & \cellcolor{blue!15}122.45$^\dagger$ & - & - \\ 
        MemVP \cite{jie2024memory} &  LLaMA   & Linear & Per-task  & 3.9M &- & - & - & - & - & \cellcolor{blue!15}\textbf{92.36} $^\dagger$ & - \\ 
        ~ & T5 base 220M & - & - & - & 7.23\% & 65.7 & - & 56 & \cellcolor{pink!30}120.8 & - & - \\ 
        ~ & ~ & ~ & ~ & ~ & ~ & \underline{59.3} & - & \underline{53.5} & \underline{116.4} & \underline{83.44} & - \\ \bottomrule
    \end{NiceTabular}
    }
    \caption{\textbf{Performance on Traditional Benchmarks.} $^\dagger$ denotes the replicated results. Others are collected from the papers or GitHub repositories. Underlined numbers are the average of each paradigm on each benchmark. Numbers in \textbf{bold} are the best results among all paradigms. Numbers in \colorbox{blue!15}{blue} background are the best results within each paradigm. Numbers in \colorbox{pink!30}{pink}  background are the second-best results within each paradigm. The LLMs are 7B unless otherwise specified.
}
\label{tab:Performance on Traditional Benchmarks}

\end{table}

\subsection{Single-stage Tuning}

Single-stage Tuning contains one pretraining stage without finetuning on downstream tasks. This training approach initially emerged during the VLPM period. Regarding parameter efficiency, pretraining a VLPM demands multiple feedforward processes, leading to a multiplicative increase in trainable parameters as the model scales up. Single-stage Tuning allows models to pretrain only an MI to connect the two modalities in a single training process. Considering the scale of LLMs, this approach is also parameter-efficient. 
For pertaining, a large number of image-text pairs are used as pretraining data; representative datasets are LAION-5B \cite{schuhmann2022laion} and 400M \cite{schuhmann2021laion}, COYO-700M \cite{kakaobrain2022coyo-700m}, Conceptual Captions 3M\cite{sharma2018conceptual} and 12M \cite{changpinyo2021conceptual}, SBU \cite{ordonez2011im2text} and Visual Genome \cite{krishna2017visual}.
ESPER \cite{ESPER2023Yu} suggests employing a training approach that does not require paired domain data, utilizing RL. It leverages the text encoder of CLIP to compute the similarity between LLM-generated text conditioned on an image and the image itself, which serves as the reward signal to align the two modalities. The RL algorithm used to optimize this reward is the clipped version of PPO \cite{schulman2017ppo,stiennon2020learning}. In the multimodal context, the reward is calculated as follows:

\begin{equation}
    \alpha \left( X_v / \|X_v\| \cdot \text{CLIP-T}(\hat{Y}) / \|\text{CLIP-T}(\hat{Y})\| \right) + \beta,
\end{equation}
where $\alpha$ and $\beta$ are fixed normalizing factors.

After pertaining, Single-stage Tuning leverages the pretrained knowledge in LLMs through zero-shot and few-shot learning, unlike VLPMs, which require end-to-end per-task fine-tuning. VLLMs are evaluated by zero-shot or few-shot Visual Question Answering (VQA) and image captioning. Few-shot prompting is a type of in-context learning \cite{schulhoff2024prompt}, which provides example question-answer pairs to the model, and no parameter updates are involved in the process. 
In most cases, only the Modality Integrator is trainable, while some models involve vision encoders. Overall, Attention-based Abstrators are mostly adopted. The representative works are Q-former \cite{li2023blip} and Resampler \cite{alayrac2022flamingo}. MAGMA and Flamingo \cite{alayrac2022flamingo} initiate the idea of utilizing both in-block and Out-of-block Integrators \cite{eichenberg2021magma}. 

 As shown in Table \ref{tab:Performance on Traditional Benchmarks}, the performance of Single-stage Tuning on traditional benchmarks highlights several key findings: \\
1. \textbf{Single-stage Tuning demonstrates limitations for downstream task generalization}, as it shows the lowest average scores among the three paradigms. This approach primarily trains the model on image-text pairs, which restricts the model's ability to generalize across various downstream tasks. The use of zero-shot learning further limits the performance, as the model lacks task-specific knowledge. \\
2. \textbf{Efficient architectural design achieves a good balance between performance and parameter efficiency}, exemplified by MAPL \cite{manas2022mapl}, which achieves the highest efficiency within the paradigm and excels in COCO captions. It further reduces the parameter of training only the MI by the efficient resampler design, projecting down and up the input and out of the original resampler structure. \\
3. \textbf{The simultaneous use of In-block and Out-of-Block Integrators proves effective}, as demonstrated by MAGMA \cite{eichenberg2021magma} and Flamingo \cite{alayrac2022flamingo}, which achieve top performance on two benchmarks each. The In-block structures are Bottleneck Adapters and Attention-based Adapters, respectively. This strategy is further applied in the Direct Adaptation paradigm.\\ 
4. \textbf{Lack of Convolution-based Abstractor.}
Designing a lightweight MI becomes crucial when it is the only tunable module in the whole training process. The advantages and drawbacks of three types of Out-of-block Integrators are summarized in Table \ref{tab: connector comparison}, inspired by \cite{cha2024honeybee}. Overall, among the Out-of-block Integrators, the linear mapping and the convolutional structure are usually more efficient than attention-based integrators. Moreover, convolutional structures have higher adaptability than Dense Integrator. However, there is a lack of initiatives to tune only the Convolution-based Abstractors.

\subsection{Two-stage Tuning} 
\label{sec: two-stage training}

Two-stage Tuning refers to the pretraining and fine-tuning process. The pretraining stage is similar to Single-stage Tuning, with the goal of aligning the visual feature with the LLM. In addition to the common datasets used in Single-stage Tuning, CC-595K filtered by Liu et al. \cite{liu2024visual} from CC3M \cite{sharma2018conceptual} and ShareGPT4V-PT \cite{chen2023sharegpt4v} with more fine-grained captions are utilized. 

The fine-tuning techniques used in Two-stage Tuning paradigm  includes per-task or multi-task learning, instruction tuning and RL, which are introduced in Section \ref{sec: Learning Paradigm}. In this training paradigm, MTL is concurrently adopted with instruction tuning, as they address different aspects of training. MTL depends on whether the training data includes multiple tasks with different output formats, while instruction tuning depends on whether instructions are provided during the training process.

\begin{table}[htb]
\fontsize{12pt}{12}\selectfont

\centering
\renewcommand{\arraystretch}{1.2} % 行高调整
\setlength{\tabcolsep}{4pt}       % 列间距调整

\resizebox{\textwidth}{!}{
\begin{tabular}{lllp{4cm}p{4cm}p{4cm}l}
\toprule
\textbf{Instruction Data} & \textbf{Size} & \textbf{Created by} & \textbf{Format} & \textbf{Source} & \textbf{Characteristics}  & \textbf{Time}\\ 
\midrule
\textbf{Text-only}               &                  &                     &                 &                 &                       &                         \\ 

ShareGPT \cite{ShareGPT}         & -                & {GPT-4/ GPT-3.5} & Multi-turn Conversation & -                 & A tool to export ChatGPT History  & {Dec 2022} \\ 
SlimOrca \cite{SlimOrca}         & 518K             & {GPT-4/ GPT-3.5} & Single turn Conversation & OpenOrca                   &    -   & Jan 2023                      \\ 
Alpaca \cite{alpaca}             & 52K              & GPT-3.5             & Single turn Conversation & -                   &   -        & { Mar 2023}               \\
Baize \cite{xu-etal-2023-baize}  & 111.5K           & GPT-3.5-turbo       & Multi-turn Conversation & -                  &  -         & {Apr 2023}               \\ 
GPT4-LLM \cite{peng2023instruction} & 52K            & GPT-4               & Single turn Conversation & Alpaca           &     -       & {Apr 2023}                \\ 
\midrule
\textbf{Multi-modal}             &                  &                     &                 &                 &                       &                         \\ 
VisDial \cite{das2017visual} &1.2M & Visual Dialog Model &Multi-turn Convernsation & COCO & The image is revealed to Visual Dialog Model when captioning   & Mar 2017\\
Video-Chat \cite{li2023videochat} & 4K/ 7K          & GPT4                & Detailed Description/ Multi-turn Conversation & WebVid-10M             & -       & April 2023                  \\ 
MiniGPT4 \cite{zhu2023minigpt}   & 3K               & -                   & Detailed Description & {CC, SBU, ALIGN} &-           & {Jun 2023}               \\ 
LRV-Instruction \cite{liu2024mitigating} & {300k} & GPT4             & Single turn Conversation & Visual Genome.   & Covering 16 vision-and-language tasks and including positive and negative instructions  & {June 2023}\\ 
LLAVA \cite{liu2024visual}       & 158K             & GPT-4               & Multi-turn Conversation & COCO                   & Context: Caption/Box; Response: conversation, detailed description, complex reasoning & July 2023     \\ 
LLAVA 1.5 \cite{liu2024improved} & 665K             & -                   & -               & LLAVA, ShareGPT, VQAv2, GQA, VG, OKVQA, OCRVQA, A-OKVQA, TextCaps, RefCOCO  &  -      & Oct 2023                  \\ 
ShareGPT4V \cite{chen2023sharegpt4v} &100K &GPT4-Vision &Detailed Description &  COCO, LAION, CC-3M, and SBU, SAM,  TextCaps,  WikiArt, webcrawled data & The image is revealed to LLMs when captioning.  & Nov 2023 \\
LLAVAR \cite{zhang2023llavar}    & 16K              & GPT-4               & Multi-turn Conversation & LAION              & -          & {Jun 2024}                \\ 
\bottomrule
\end{tabular}
}
\caption{\textbf{Summary of Instruction Tuning Data.}}
\label{tab:instruction_tuning}

\end{table}

\begin{table}[!htb]
\fontsize{13pt}{15}\selectfont

    \centering
    \renewcommand{\arraystretch}{1.2} % 行高调整
    \setlength{\tabcolsep}{4pt}       % 列间距调整
    
    \resizebox{\textwidth}{!}{
    \begin{NiceTabular}{llp{5cm}llllllllll}
     \toprule
    \textbf{Model} & \textbf{LLM} & \textbf{Trainable} & \textbf{Para \#} & \textbf{PT/FT Size} & \textbf{POPE} & \textbf{MM-Bench} & \textbf{MM-Vet} & \textbf{MME} & \textbf{MME\_P} & \textbf{SEED} & \textbf{SEED\_I} & \textbf{LLaVA-Bench-Wild} \\ 
    \midrule
    LLaMA-Adapter \cite{zhang2023llama} & LLaMA 7B & Linear + Adaptive Prefix & 1.8M & - & - & - & - & - &\cellcolor{blue!15} 973 & - & - & - \\ 
    LLaVA \cite{liu2024visual} & LLaMA-2-7B-Chat & Linear + LoRA & 4.4M & 558K/158K & - & - & - & - & - &-& - & 62.8 \\ 
    MiniGPT4 \cite{zhu2023minigpt} & Vicuna & Linear & - & 558K/665K & - & 24.3 & 22.1 & 726 & 867 & - & \cellcolor{blue!15}47.4 & - \\ 
    mPLUG-Owl1 \cite{ye2023mplug} & LLaMA & Resampler + LoRA & - & 2.1M/102K & - & 46.6 & - & 967.34 & - & 34 & - & - \\ 
    LLaMA-AdapterV2  \cite{gao2023llama} & LLaMA & Linear Projector + Degree-Adaptive Prefix + Bias, Norm & 14M & 567K/52K & - & 41 & 31.4 & 1221.6 & 972.7 & - & 32.7 & - \\ 
    LLava 1.5 \cite{liu2024improved} & Vicuna v1.5 7B & MLP+LoRA & - & 558K/665K & \cellcolor{pink!30}86.4 & \cellcolor{pink!30}66.1 & 30.2 & \cellcolor{blue!15}1476.9 & - & \cellcolor{pink!30}60.1 & -& \cellcolor{pink!30}67.9 \\

    CogVLM-Chat \cite{wang2023cogvlm} & Vicuna-v1.5 & MI+VE & - & 1.5B/6M & \cellcolor{blue!15}87.9 & \cellcolor{blue!15}77.6 &\cellcolor{blue!15} 51.1 & - & - & \cellcolor{blue!15}72.5 & - & \cellcolor{blue!15}77.8 \\ 
    MobileVLM \cite{chu2023mobilevlm} & MobileLLaMA 2.7B & Convolution + Small LLM & 2.71B & 558K/665K & 84.9 & 59.6 & - & 1288.9 & - & - & -& - \\ 
    ~ & ~ & Convolution + LoRA & 201.71 M & 558K/665K & 84.6 & 57.0 & - & 1296.4 & - & -& - & - \\ 
    MobileVLM v2 \cite{chu2024mobilevlm} & MobileLLaMA 2.7B & Convolution + Small LLM & 2.7B & 1.2M/2.4M & 84.7 & 63.2 & - &\cellcolor{pink!30} 1440.5 & - & - & - & - \\ 
    VL-Mamba \cite{qiao2024vl} & Mamba LLM-2.8B & VSS +Small LLM & - & 558K/665K & 84.4 & 57 &\cellcolor{pink!30} 32.6 & 1369.6 & - & - & - & - \\ 
    \bottomrule
    \end{NiceTabular}
    }
    \caption{\textbf{Performance of Two-stage Tuning on Advanced Benchmarks.}}
    \label{tab: Advanced Benchmarks}
\end{table}

Instruction Tuning is adopted to equip VLLM with the ability to understand user intentions and improve generalization ability \cite{wei2022finetuned,liu2024visual}. The instruction data is built by organizing structured datasets in natural language, the idea of which is brought to the multimodal domain by LLaVA \cite{liu2024visual}. LLaVA \cite{liu2024visual} uses image captions and bounding boxes to express visual features to prompt text-only GPT4 to generate three types of instruction data: conversation, detailed description, and complex reasoning. The datasets for instruction tuning are summarized in Table \ref{tab:instruction_tuning}. Another approach to instruction tuning involves using instruction prompts to format the responses generated by the VLLM. In this method, the training data typically consists of detailed descriptions \cite{zhu2023minigpt, liu2024improved, chen2023sharegpt4v}, enabling the model to produce outputs that align more closely with the provided instructions.

For RL, Sun et al. \cite{sun2024llavaRLHF} applied RLHF in the multimodal domain to eliminate hallucinations of images by language models. RLAIF \cite{RLAIF2024Ahn} applies RL with AI feedback. It uses video instruction data to tune VLLMs, generating responses and preferences to train a reward model.

There are three ways to save computational resources under the Two-stage Tuning setting. Like the other two paradigms,  tuning only the Modality Integrator is the most efficient way. Among the works \cite{zhu2023minigpt, zhang2023video, wang2023cogvlm,alayrac2022flamingo, li2023blip,zhu2023minigpt}, Mini-GPT4 only tunes a single linear layer. For those tuning LLMs in the second stage, LoRA-based Tuning discussed in Section \ref{sec: LoRA} is an option. mPLUG-OWL1 \cite{ye2023mplug} proposes to employ LoRA \cite{hu2021lora}, and Multimodal-GPT \cite{gong2023multimodal} finetunes openFlamingo \cite{awadalla2023openflamingo} by adding LoRA into the attention module and FFN module. Inspiring work DoRA \cite{liu2024dora} shows superior performance over full-finetuning and LoRA \cite{hu2021lora} in LLaVA v1.5, with tuning 4.63\% of the parameters. The other way is to employ small-scale LLMs, as in MobileVLM \cite{chu2023mobilevlm, chu2024mobilevlm} and VL-Mamba \cite{qiao2024vl}.

The availability of code implementations is summarized in Table \ref{tab:Code implementations}. Table \ref{tab:Performance on Traditional Benchmarks} and Table \ref{tab: Advanced Benchmarks} reveal several important insights regarding the performance of Two-stage Tuning on traditional and advanced benchmarks:\\
1. \textbf{The additional stage of tuning improves effectiveness, especially Instruction Tuning.}
Overall, Two-stage tuning outperforms the other two paradigms, yielding the best results across five benchmarks. This paradigm achieves the top results in most of the benchmarks, with LLaVA 1.5 \cite{liu2024improved}, utilizing 665K instructional data, securing a top-2 performance ranking. On more advanced benchmarks, both LLaVA 1.5 \cite{liu2024improved} and COGVLM-chat  \cite{wang2023cogvlm} demonstrate competitive performance, consistently producing top-2 results across the majority of advanced benchmarks.\\
2. \textbf{Training only the MI can be effective, and non-updated LLMs can also achieve good performance.}
COGVLM-chat \cite{wang2023cogvlm} achieves strong results across three VQA benchmarks by training only the linear adapters and the vision encoder without requiring updates to the LLM. Similarly, BLIP2 \cite{li2023blip} shows notable performance across two benchmarks, highlighting the effectiveness of the Q-former. LlaMA-Adapter \cite{zhang2023llama} tunes 1.8M parameters while achieving the best results on the MME perception set, demonstrating its comparable capabilities in perception tasks.\\
3. \textbf{Reparameterization techniques demonstrate effectiveness.}
LLaVA 1.5 \cite{liu2024improved}, utilizing LoRA and Dora, and exhibits robust performance in GQA, outperforming models that tune the LLM. This illustrates the effectiveness of reparameterization techniques in enhancing the performance of the modality integrator without requiring extensive updates to the base model.\\
4. \textbf{Leveraging small-scale LLMs represents a new approach to saving parameters while maintaining competitive performance.}
MobileVLM \cite{chu2023mobilevlm} and VL-Mamba \cite{gu2023mamba} achieve the best scores on MM-VET and MME, respectively, highlighting the potential of small-scale LLMs in achieving competitive performance on multimodal benchmarks.

\begin{table}[htb]
\centering

\resizebox{\textwidth}{!}{%
\begin{NiceTabular}{lccccccc p{5cm}}
\toprule
\multirow{2}{*}{\textbf{Model}} & \multicolumn{2}{c}{\textbf{Pretraining}}         & \multicolumn{4}{c}{\textbf{Finetuning}}  & \multicolumn{2}{c}{\textbf{Evaluation}} \\
 &  \textbf{Code} & \textbf{Data}   & \textbf{Code} & \textbf{LoRA} & \textbf{Data} & \textbf{Instruction Template} &\textbf{UI for Inference} & \textbf{Supported Datasets}  \\ \midrule
OpenFlamingo \cite{awadalla2023openflamingo}         & \tick         & \cross        & \tick             & -       & \cross        & -        & \cross                             & COCO, Flickr-30K, VQA v2, \\ 
&  & &  & &  &   &  & OK-VQA, TextVQA, VizWiz, \\ 
 &  & &  & &  &   &  &  Hateful Memes, ImageNet \\ \midrule
LLaVA \cite{liu2024visual}   & \tick   & \tick & \tick  & LoRA/qLoRA  & \tick & \tick         & \tick    & VQAv2, GQA, VisWiz, \\ 
&  & &  & &  &   &  & ScienceQA, TextVQA, POPE, \\ 
 &  & &  & &  &   &  &MME, MM-Bench, MMVet, \\ 
&  & &  & &  &   &  & LLaVA-Bench-in-the-Wild \\ \midrule
LLaVA 1.5 \cite{liu2024improved}                    & \tick         & \tick              & \tick     & LoRA        & \tick        & \tick         & \tick                               & Same as LLaVA \\\midrule
MiniGPT4 \cite{zhu2023minigpt}                      & \tick         & \tick  & \tick       & -          & \tick          & \tick       & \cross     & RefCOCO, RefCOCO+, \\ 
&  & &  & &  &   &  & RefCOCOg,OKVQA,VIZWIZ,\\
&  & &  & &  &   &  &ICONVQA,GQA,VSR,HM \\ \midrule
mPLUG-Owl 1 \cite{ye2023mplug}                     & \cross         & \cross                 & \tick    & LoRA      & \cross          & \tick         & \tick                             & OwlEval \\ \midrule
mPLUG-Owl 2 \cite{ye2024mplug}                      & \cross         & \cross                 & \tick    & LoRA      & \cross        & \tick         & \tick                                & Flickr 30k, COCO, VQA v2, \\ 
&  & &  & &  &   &  & OKVQA, TextVQA, MM-Bench \\ \midrule
LLaMA-Adapter v2.1 \cite{gao2023llama}              & \tick         & \cross & \tick        & -          & \tick         & -                               & \tick      & MME \\ \midrule
Video-LLaMA \cite{zhang2023video}                  & \tick         & \tick               & \tick         & -  & \tick          & -       & \cross     &  \cross   \\ \midrule
BLIP-2 \cite{li2023blip}                            & \tick         & \tick                  & \tick         & -          & \tick        & -                            & \cross     & Common datasets \\\midrule
QWEN-VL \cite{bai2023qwen} & \cross         & \cross    & \tick       & LoRA/qLoRA   & \cross        & \tick        & \tick         & \cross                         \\ \midrule
COGVLM \cite{wang2023cogvlm}                        & \cross        & \cross        & \tick           & -     & \cross    & -      & \tick   & Captcha Images dataset \\\midrule
Honeybee \cite{cha2024honeybee}     & \tick     & \tick         & \tick      & \cross            & \tick         & \tick                  & \cross    & MMB, MME, SEED-Bench, \\ 
&  & &  & &  &   &  & ScienceQA,LLaVABench,MMVet, \\
&  & &  & &  &   &  & MMMU,POPE, OwlEval\\ \midrule
MobileVLM v1/v2 \cite{chu2023mobilevlm} \cite{chu2024mobilevlm}            & \tick         & \tick   & \tick       & LoRA         & \tick         & -          & \cross     & GQA, MMBench, \\ 
                              &  & &  & &  &   &  & MME, POPE, SQA, \\ 
                              &  & &  & &  &   &  & TextVQA \\ \bottomrule
\end{NiceTabular}%
}
\caption{\textbf{Code Availability of VLLMs with Two-stage Tuning.} "-" denotes that the column is not applicable to the model.  }
\label{tab:Code implementations}
\end{table}

\subsection{Direct Adaptation} 

Direct Adaptation skips the pretraining process and directly fintunes the model to specific downstream tasks. Although Single-stage Tuning and Direct Adaptation both involve one training phase, the knowledge learned from training is different. Taking the VQA task as an example, for Single-stage Tuning, the model learns from descriptive text associated with the images during pretraining. However, during testing, the models are required to answer questions about seen or new images. For Direct Adaptation, the models are directly trained on the data of QA format but tested on unseen questions. 

For the learning paradigm, VL-adapter \cite{sung2022vl} was the first to propose the use of MTL, followed by LST \cite{sung2022lst} and VL-PET \cite{hu2023vl}. VL-PET highlights the importance of MTL in enhancing model performance and minimizing storage requirements. LaVIN \cite{luo2024cheap} was the pioneer in applying instruction tuning to the direct adaptation paradigm. 

With reducing parameters as the primary goal, VLLMs adopting Direct Adaptation only train the Modality Integrator. Usually, these models utilize both out-of-block and In-block Integrators. A common measure is to first linearly map the visual embedding to the LLM input dimension and then fuse the visual information with In-block Integrators, which are efficient structures, as discussed in Section \ref{sec: In-block Integrator}. Besides, LLMs of million-level sizes are more often adopted, such as T5 and Bart serving as benchmark LLMs in VL-Adapter \cite{lin2020exploring}, VL-PET \cite{hu2023vl}, LaVIN \cite{luo2024cheap}, MemVP \cite{jie2024memory}, LST \cite{sung2022lst}.

\begin{table*}[!htb]  

\centering  
\resizebox{\textwidth}{!}{
\begin{NiceTabular}{lccccccccc}  
\toprule
\multirow{2}{*}{Method}  &\multirow{2}{*}{\makecell[c]{LLM}} & \multirow{2}{*}{\makecell[c]{Trainable \\Params \%} }&\multirow{2}{*}{\makecell[c]{Tunable \\Component}} &\multirow{2}{*}{\makecell[c]{Peak \\Memory}} &\multirow{2}{*}{\makecell[c]{VQAv2}}  &\multirow{2}{*}{\makecell[c]{GQA} }  &\multirow{2}{*}{\makecell[c]{COCO \\Captions}}&\multirow{2}{*}{\makecell[c]{NLVR} }   &  \multirow{2}{*}{Average}  \\ 
\\
\midrule

\textbf{Direct Adaptation} &&&&&&&&&\\ 
Full Fine-Tuning$^\dagger$ &T5$_{base}$ 220M &100\% &Linear + LLM &37.78G &\cellcolor{pink!30}67.13 &\cellcolor{blue!15}{56.46}  &112.42 &\cellcolor{pink!30}73.60 &77.40\\
Prompt Tuning$^\dagger$ &T5$_{base}$ 220M &\cellcolor{blue!15}1.26\% &Linear + Prompt  &40.03G &47.64 &41.12 &96.64 &51.17 &59.14\\
LoRA$^\dagger$ &T5$_{base}$ 220M &7.54\% &Linear + LoRA  &29.52G &63.88 & 52.88 &110.75 &70.32 &74.46 \\ \hdashline

VL-Adapter$^\dagger$ &T5$_{base}$ 220M  &7.98\% &Linear + Bottleneck Adapter  &30.06G &66.99 &56.36 &111.85 &73.07 &77.07 \\
VL-PET$_{Small}^\dagger$ &T5$_{base}$ 220M  & 4.51 \% &Linear + Bottleneck Adapter &- &65.7 &55.78 &119.46 &73.45 &78.60 \\
VL-PET$_{MiddleX}^\dagger$ &T5$_{base}$ 220M &\cellcolor{pink!30} 4.50 \% &Linear + Bottleneck Adapter &- &66.54 &56.38 &120.10 &\cellcolor{blue!15}{74.72} &\cellcolor{pink!30}79.43 \\
VL-PET$_{MiddleY}^\dagger$ &T5$_{base}$ 220M & \cellcolor{pink!30} 4.50 \% &Linear + Bottleneck Adapter  &- &65.03 &55.71 &118.07 &72.14 &77.74 \\ % 20 epocs （时间长）
%test wandb_log_dict =  {'VQA/Test/overall': 65.03, 'VQA/Test/topk_optimal': 68.49, 'VQA/Test/topk_not_optimal': 10.03, 'NLVR/Test/accuracy': 72.14008899095738, 'NLVR/Test/consistency': 35.58897243107769, 'Caption/Test/Bleu_1': 0.7677807214183663, 'Caption/Test/Bleu_2': 0.6119640398782643, 'Caption/Test/Bleu_3': 0.46838293106890866, 'Caption/Test/Bleu_4': 0.3530112000762595, 'Caption/Test/METEOR': 0.28205845105196586, 'Caption/Test/ROUGE_L': 0.5722621517943879, 'Caption/Test/CIDEr': 1.1806849143859275, 'Caption/Test/SPICE': 0.218860804141495}
VL-PET$_{Large}^\dagger$ &T5$_{base}$ 220M & 7.31\% &Linear + Bottleneck Adapter &- &67.04 &55.97 &\cellcolor{blue!15}{122.45}  &72.93 &\cellcolor{blue!15}{79.60} \\

LST$^\dagger$ &T5$_{base}$ 220M &7.46\% &Linear + Attention-based Adapter  &15.04G & \cellcolor{blue!15}{67.22} &\cellcolor{pink!30}56.37 &115.05 &73.04 &77.92 \\
MemVP &T5$_{base}$ 220M &7.23\% $^3$ &Linear &- &65.7 &56.0 &\cellcolor{pink!30}120.8 &- &-\\

\bottomrule
\end{NiceTabular}  
}
\caption{\textbf{Performance of Direct Adaptation on VQAv2, GQA, and COCO Captions.}  $^\dagger$ denotes the replicated results. The result of MemVP \cite{jie2024memory} is from the paper, which adopts the per-task adaptation and shows the average result of three runs. Others are multi-task learning and the result of one run. The MemVP trainable parameter \% is computed based on the proportion to VL-PET$_{Large}$.}  
\label{tab: DA experiment}

\end{table*} 

Most of the results for Direct Adaptation are reproduced as indicated in Table \ref{tab:Performance on Traditional Benchmarks} and Table \ref{tab: DA experiment}.  The reproduced results are a single-run outcome. The performance for LaVIN \cite{luo2024cheap} and MemVP \cite{jie2024memory} in Table \ref{tab:Performance on Traditional Benchmarks} and VL-PET \cite{hu2023vl} and LST \cite{sung2022lst} in Table \ref{tab: DA experiment} are reproduced using the official code repositories. For other methods, reproductions are based on the LST repository \cite{sung2022lst}. All experiments are run on A100 GPUs. The performance of Direct Adaptation on traditional benchmarks highlights key observations: \\
1. \textbf{Supervised multi-task fine-tuning proves effective}, as evidenced by significant performance improvements over Single-stage Tuning. Compared to Single-Stage Tuning, despite both involving one training stage, Direct Adaptation demonstrates significant improvements, with average score increases of 7.8, 10.5, and 22.7 on VQAv2, GQA, and COCO benchmarks, respectively. MemVP  \cite{jie2024memory} achieves the best performance on the SQA benchmark among the three paradigms, highlighting its focus on parameter efficiency while maintaining competitive performance. \\
2. \textbf{MI in Direct Adaptation focuses on parameter efficiency while retaining good performance}. The idea of bottleneck adapters has mainly been adopted.  As shown in Table \ref{tab: DA experiment}, most MI trains fewer parameters than LoRA, among which the VL-PET series \cite{hu2023vl} is the most parameter-efficient, with VL-PET$_{Large}$ achieving the best average result. Notably, most models outperform full fine-tuning in terms of average performance. \\
3. \textbf{Memory efficiency also plays a crucial role}. LST \cite{sung2022lst} demonstrates its effectiveness by reducing memory usage by over 50\%, showcasing the potential for scalable applications. \\
4. \textbf{A trend has emerged to adapt larger-scale LLMs and adopt Instruction Tuning}. In the Direct Adaptation framework, T5 \cite{raffel2020exploring} and Bart \cite{chung2024scaling} are utilized as the standard LLMs because of their practical sizes \cite{sung2022vl, sung2022lst, hu2023vl, luo2024cheap, jie2024memory}. Furthermore, some attempts are to adapt LLaMA models within this framework \cite{luo2024cheap,jie2024memory, zhang2023llama,gao2023llama}. As Instruction Tuning is effective in Two-stage Tuning, LaVIN \cite{luo2024cheap} instructionally adapts LLaMA without pretraining, outperforming LLaVA1.5 on the SQA benchmark.

\section{Conclusion}
\label{sec: future direction}
In this survey, we have investigated the evolution of training paradigms and methodology of integrating vision modalities into LLMs to create VLLMs, focusing on parameter efficiency. We categorized the training paradigms into three types: Single-stage Tuning, Two-stage Tuning, and Direct Adaptation. Each paradigm offers unique advantages and efficiency strategies. Single-stage Tuning, which emerged during the VLPM era, effectively leverages pretrained knowledge in LLMs with small additional training, primarily through a Modality Integrator. However, it does not fully unleash the instruction-following potential of LLMs. Two-stage Tuning introduces an additional phase to enhance zero-shot transfer and user intention understanding, focusing on multi-task learning and instruction tuning. This paradigm employs various strategies to reduce trainable parameters, including selective training of the MI, the incorporation of reparameterization modules, and the employment of small-size LLMs. Despite its resource-intensive nature, this approach significantly improves LLMs' generalization and reasoning abilities. Direct Adaptation, on the other hand, aims to minimize resource consumption by directly fine-tuning the MI on downstream tasks without pretraining, thus providing a balance between parameter efficiency and performance.

In summary, our review highlights several key takeaways that suggest future directions for research in integrating vision modalities into LLMs: among the evaluated paradigms, Two-stage Tuning demonstrated the highest performance, followed by Direct Adaptation, with Single-stage Tuning showing the lowest performance. Meanwhile, Direct Adaptation primarily focuses on parameter efficiency, indicating that the paradigm shift from Single-stage Tuning to the other two methods holds significant potential.
Secondly, instruction tuning has emerged as the most popular fine-tuning technique within the Two-stage Tuning paradigm, and there are new attempts to incorporate it into Direct Adaptation. Thirdly, the efficient design of the MI is crucial across all three paradigms. Direct Adaptation, in particular, has led the way in achieving efficiency. Applying efficient MI designs to larger LLMs is becoming increasingly necessary.

\section*{Acknowledgment}
The research described in this article has been supported by a grant from the Research Grants Council of the Hong Kong Special Administrative Region, China (R1015-23) and the Faculty Research Grant (SDS24A8) and the Direct Grant (DR25E8) of Lingnan University, Hong Kong.

\newcommand{\cmark}{\ding{51}}% check mark
\newcommand{\xmark}{\ding{55}}% cross mark

%Bibliography
\bibliographystyle{unsrt}  
\bibliography{main}

\end{document}